\documentclass[preprint,12pt]{elsarticle}

\usepackage{amssymb,}
\usepackage{amsmath}
\usepackage{graphicx,tabularx,booktabs, subcaption}

\journal{Robotics and Autonomous Systems}

\begin{document}

\begin{frontmatter}



\title{Offline Reinforcement Learning using Human-Aligned Reward Labeling for Autonomous Emergency Braking in Occluded Pedestrian Crossing}


\author[1]{Vinal~Asodia\corref{cor1}}
\ead{va00191@surrey.ac.uk}
\cortext[cor1]{Corresponding author}

\author[1]{Barkin~Dagda}
\ead{bd00242@surrey.ac.uk}

\author[1]{Yinglong~He}
\ead{yinglong.he@surrey.ac.uk}

\author[2]{Zhenhua~Feng}
\ead{fengzhenhua@jiangnan.edu.cn}

\author[1]{Saber~Fallah}
\ead{s.fallah@surrey.ac.uk}

\affiliation[1]{organization={CAV-Lab, School of Engineering},
            addressline={University of Surrey},
            city={Guildford},
            postcode={GU2 7XH},
            country={United Kingdom}}

\affiliation[2]{organization={School of Artificial Intelligence and Computer Science},
            addressline={Jiangnan University},
            city={Wuxi},
            postcode={213122},
            country={China}}

\begin{abstract}
Effective leveraging of real-world driving datasets is crucial for enhancing the training of autonomous driving systems. While Offline Reinforcement Learning enables training autonomous vehicles with such data, most available datasets lack meaningful reward labels. Reward labeling is essential as it provides feedback for the learning algorithm to distinguish between desirable and undesirable behaviors, thereby improving policy performance. This paper presents a novel approach for generating human-aligned reward labels. The proposed approach addresses the challenge of absent reward signals in the real-world datasets by generating labels that reflect human judgment and safety considerations. The reward function incorporates an adaptive safety component that is activated by analyzing semantic segmentation maps, enabling the autonomous vehicle to prioritize safety over efficiency in potential collision scenarios. The proposed method is applied to an occluded pedestrian crossing scenario with varying pedestrian traffic levels, using simulation data. When the generated rewards were used to train various Offline Reinforcement Learning algorithms, each model produced a meaningful policy, demonstrating the method's viability. In addition, the method was applied to a subset of the Audi Autonomous Driving Dataset, and the reward labels were compared to human-annotated reward labels. The findings show a moderate disparity between the two reward sets, and, most interestingly, the method flagged unsafe states that the human annotator missed. 
\end{abstract}


\begin{highlights}
\item Proposes a reward labeling pipeline to enable autonomous driving datasets to be used for training Offline Reinforcement Learning agents.
\item Introduces a reward function that utilizes an adaptive safety component that uses semantic segmentation maps to assess critical risk factors within the driving scene.
\item Presents a method to use semantic segmentation maps to apply varying levels of spatial attention to different objects in the image observation.
\end{highlights}


\end{frontmatter}



\section{Introduction}
\label{section:introduction}

Autonomous Driving (AD) simulators have played a crucial role in the development and evaluation of Autonomous Vehicle (AV) systems under safe, controlled, and reproducible conditions. These simulation platforms facilitate scalable data collection, as well as the generation of diverse and complex traffic scenarios, without the risks associated with real-world experimentation. Within these environments, Imitation Learning (IL) has emerged as an effective approach for training driving policies by mimicking expert demonstrations~\cite{kuutti2021adversarial}, \cite{arbabi2022learning}, \cite{arbabi2022planning}. However, IL agents often suffer from limited generalization when exposed to situations out-of-distribution (OOD) from the training dataset. To address this limitation, researchers have adopted Reinforcement Learning (RL), which trains agents via continuous online interactions with the simulated environment. Noteworthy recent examples include~\cite{kuutti2021arc}, \cite{liu2024augmenting}, \cite{wang2024explainable}, where RL was utilized to train AVs to complete tasks like navigating through highways, busy intersections, and roundabouts.

However, it is well documented that there is a performance degradation of an RL policy trained within a simulated environment and deployed in the real world~\cite{salvato2021crossing}, \cite{zhao2020sim}, \cite{stocco2022mind}. This is known as the "sim-to-real" gap, and contributing factors could be environmental complexity (i.e., real-world variability), simulation sensor limitations (i.e., sensor fidelity and calibration), and differences in vehicle dynamics. There are ongoing efforts within the research community to bridge the sim-to-real gap for AD. Such avenues include; utilizing digital twins to pretrain AVs on synthetic data \cite{voogd2023reinforcement}, guiding the RL system during training and deployment with human-behavior samples \cite{wu2023human}, building realistic simulators with real-world data \cite{gulino2024waymax}, and applying methods like domain randomization to make the system more robust to stochastic changes in the environment \cite{toth2024sim}. Although there are strong efforts to close the sim-to-real gap, an alternative solution is to avoid interacting with simulated environments during training, thus avoiding the challenge outright.

In recent years, the AD research community has shown a growing interest in using datasets collected in the real world to train AVs, rather than relying on interactions within simulated environments. Offline RL \cite{levine2020offline}, a data-driven learning paradigm that aims to train a policy directly from a static dataset. To date, the research on Offline RL has focused primarily on addressing the distributional shift that occurs when the agent encounters unseen states during deployment (i.e., OOD samples), and previous researchers have mainly evaluated their pipelines on robotic tasks~\cite{sinha2022s4rl}, \cite{bhardwaj2024adversarial}, \cite{hong2024learning}. There is a limited number of studies that address the challenge of OOD samples within the AD domain, with few previous works tackling lane change and parking scenarios \cite{shi2021offline}, as well as roundabout and various highway scenarios \cite{lin2024safety}. The aforementioned studies have the shared limitation of being conducted in relatively low-fidelity simulation platforms that have a wide sim-to-real gap, raising questions on the validity of said approaches for the real world. Recent studies have addressed this limitation, tackling navigating through intersections \cite{diehl2023uncertainty}, and combinations of driving scenarios \cite{li2024boosting} in the high-fidelity urban driving simulator, CARLA \cite{dosovitskiy2017carla}. Overall, these studies show great promise in handling a number of driving scenarios, however, there are also a number of complex scenarios (e.g., occluded pedestrian crossings) that lack attention in the Offline RL research space. 

To promote further research within the Offline RL community for AD, researchers have created datasets with real-world driving data for Offline RL~\cite{fang2022offline}, \cite{lee2024ad4rl}. While these datasets offer an entry point for future Offline RL work, a common theme across previous work is the use of a low-dimensional vector-state representation. Consequently, these datasets are unsuitable for Offline RL pipelines that use image-state observations, creating a gap in the perception-based Offline RL research space. Meanwhile, a variety of extensive real-world datasets, such as the Waymo Open Dataset \cite{sun2020scalability}, nuScenes \cite{nuscenes}, and the Audi Autonomous Driving Dataset (A2D2) \cite{geyer2020a2d2} exist. However, these datasets lack meaningful reward labels, which prevents them from being used for Offline RL training.

To allow existing datasets to be used for Offline RL training, they must be put through a reward labeling process, which involves inferring or designing reward functions to guide policy learning. In particular for AD tasks, these rewards need to align with human values (e.g., legality, fairness, and safety) if the agent is to be deployed in the real world and coexist with other human drivers. One of the earlier methods, proposed by Yu et al. \cite{yu2022leverage}, asserted that if one has access to a labeled dataset, then one can simply use a zero reward label to leverage new unlabeled data for policy learning. A key limitation is that the approach was primarily tested in closed environment tasks (robotic locomotion and manipulation), and raises safety concerns if applied to more complex AD tasks. Later, Sun et al. proposed to use reward machines to encode task related human knowledge and label offline datasets \cite{sun2023less}. The authors' main focus was towards long horizon tasks, where they produce counterfactual experiences to refine the dataset, resulting in a better end policy. Recently, researchers have proposed an alternative approach to reward labeling, which involves using human preferences between different trajectories to supply binary reward labels~\cite{xu2024binary}, \cite{venkataraman2024real}. Overall, despite the importance of reward labeling for Offline RL, very little research has been conducted within the AD domain, as the research community predominantly relies on interacting with simulation platforms for their works. Specifically, less focus has been given on plethora of complex AD scenarios, where relying on human preferences or relatively simplistic reward labeling methods may not be sufficient. 

In summary, the key contributions of this paper include:
\begin{itemize}
    \item We introduce a novel approach for generating human-aligned reward labels for end-to-end Offline RL by utilizing the existing AD datasets.
    \item We introduce a reward function that incorporates an adaptive safety mechanism using semantic segmentation maps to assess critical risk factors, such as pedestrian locations and zebra crossing presence, enabling enhanced situational awareness through the tracking of multiple object classes.
    \item We leverage semantic maps to provide varying levels of spatial attention to different object classes, allowing AVs to focus on salient elements in the driving scene and significantly improve decision-making.
\end{itemize}

The rest of the paper is structured as follows. Section \ref{section:problem_formulation} details the problem formulation, while Section \ref{section:reward_labeling} breaks down the proposed reward labeling method. Section \ref{section:offline_rl} then outlines the Offline RL training procedure. Section \ref{section:experimental_setup} describes the experimental setup, including both simulation and real-world data used to evaluate the proposed method. Section \ref{section:results} presents the experimental results, followed by a discussion and potential future directions in Section \ref{section:discussion}. Finally, conclusions are drawn in Section \ref{section:conclusion}.

\section{Problem Formulation}
\label{section:problem_formulation}
This paper proposes a reward labeling method for AD datasets to facilitate Offline RL training. In particular, it focuses on the vehicle longitudinal control task under the occluded pedestrian crossing scenario (defined in Section \ref{section:scenario}). This section explains how the task is formulated as a Markov Decision Process (MDP), followed by the state, action, and reward definitions. 

\subsection{MDP Formulation}
\label{section:problem_formulation_mdp}
Within RL, an agent learns to complete a given task via interactions with the environment. It is common for RL problems to be formulated as a Markov Decision Process (MDP), giving the tuple $\{\mathcal{S}, \mathcal{A}, \mathcal{P}, \mathcal{R}, \gamma\}$, where:
\begin{itemize}
    \item $\mathcal{S}$: state set, denotes the agent's observation of the environment.
    \item $\mathcal{A}$: action set, denotes the list of available actions the agent may take.
    \item $\mathcal{P}$: transition dynamics, denotes the transition probability function $\mathcal{P}: \mathcal{S} \times \mathcal{A} \rightarrow \mathcal{S}$, which dictates the next state the agent will be in, given the current state and action.
    \item $\mathcal{R}$: reward function, denotes the function $\mathcal{R}(s, a, s')$, which gives a numerical reward to the agent for transitioning from state $s$ to state $s'$ after taking action $a$.
    \item $\gamma$: discount factor $\gamma \in [0, 1]$, denotes the agent's weighting of short-term and long-term rewards.
\end{itemize}

The primary objective in RL is to learn a policy, $\pi : \mathcal{S} \rightarrow \mathcal{A}$, that maximizes the expected cumulative reward over time. Offline RL extends RL, where the agent does not interact with the environment during training. Instead the agent learns a policy $\pi$ directly from a fixed dataset $\mathcal{D} = \{(s_{i}, a_{i}, r_{i}, s'_{i})\}^{\mathcal{N}}_{i=1}$ that is comprised of $\mathcal{N}$ tuples, collected with one or more policies (e.g., a behavior policy $\pi_b$, random policy $\pi_r$ etc.). The following sub-sections define the state space, action space, and the reward function for the vehicle longitudinal control task under the occluded pedestrian crossing scenario (full scenario description given in Section \ref{section:scenario}). 

\subsubsection{State Space ($\mathcal{S}$)}
\label{section:method_state_space}
The observation space comprises $1\times224\times224$ grayscale images captured by a front facing camera mounted on the ego vehicle. The grayscale format has been chosen to reduce the system's computational load. More details are provided in Section \ref{section:method_spatial_attention} on how the camera images are processed to provide the necessary information to the Offline RL agent.

\subsubsection{Action Space ($\mathcal{A}$)}
\label{section:method_action_space}
To ensure smooth and efficient longitudinal control, the Offline RL agent will output a continuous action within the range of $[-1, 1]$ to manage the speed of the ego vehicle. If the action is greater than $0$, the agent will increase the vehicle's throttle. Conversely, if the action is less than $0$, the agent will apply the corresponding level of braking. This continuous action output enables the agent to smoothly regulate the ego vehicle's speed.
\subsubsection{Reward Function ($\mathcal{R}$)}
\label{section:method_reward_function}
The reward function from \cite{thornton2018value} has been adapted, as it considers human values like safety and efficiency, which are associated with the ego vehicle's longitudinal control. The reward function is divided into three components: safety (Equation \ref{eqn:safe}), efficiency (Equation \ref{eqn:efficient}), and smoothness (Equation \ref{eqn:smooth}). Each reward component takes in the current state $x_t$ and action $u_t$ as input. The safety component is as follows:
\begin{equation}
    g_{safe}(x_{t}, u_{t}) = -(\zeta \frac{v_{t}^2}{d_{t}+\epsilon} + \eta \bold{1}(d_{t} = 0))\bold{1}(c_{t}),
    \label{eqn:safe}
\end{equation}
\noindent where $\zeta > 0$ is a penalty weight applied if the ego vehicle drives too fast near a pedestrian, $v_t$ is the ego vehicle’s speed, $d_t$ is the distance between the ego vehicle and the pedestrian, $\epsilon > 0$ is a constant to offset the denominator, $\eta > 0$ is a penalty to dissuade the ego vehicle from colliding with the pedestrian. This penalty component is multiplied by the indicator function $\bold{1}(d_t=0)$, that will turn the component on ($\bold{1}$) if the distance between the vehicle and the pedestrian is $0$, otherwise the penalty component is turned off ($\bold{0}$). 
Finally, there is an additional indicator function $\bold{1}(c_t)$ that will turn the entire component on if the adaptive safety flag is true ($c_t=1$), otherwise the component is turned off. Further details on $c_t$ are provided in Section \ref{section:method_adaptive_safety_component}. Altogether the reward component $g_{safe}$ encapsulates the value of safety, penalizing the agent for driving towards a pedestrian and for any collisions. 
The second component ensures efficiency in the vehicle’s control and is formulated as:
\begin{equation}
    g_{efficient}(x_{t}, u_{t}) = \mu v_{t} \bold{1}(\neg c_{t}),
    \label{eqn:efficient}
\end{equation}
\noindent where $\mu > 0$ is a weight on the reward term that encourages the vehicle to move faster when no collision is imminent. Similar to the safety component, an indicator function $\bold{1}(\neg c_t)$ is used to switch on the efficiency component if the negation of the adaptive safety flag is true (i.e., $c_t=0$), otherwise the component is turned off.
To ensure smooth and comfortable control for passengers, the final component penalizes large acceleration values $a_t$, which would cause jerky motions. The smoothness component is defined in Equation \ref{eqn:smooth}, where $\xi > 0$ is a weight to tune the penalty. It should be noted that this component is on permanently, thus $c_t$ is not used.
\begin{equation}
    g_{smooth}(x_{t}, u_{t}) = -\xi(v_{t} - v_{t+1})^2 = -\xi(a_{t}\Delta t)^2 , 
    \label{eqn:smooth}
\end{equation}

The final reward value is the summation of the aforementioned components (Equation \ref{eqn:final_reward}) and allows the RL agent to shift in driving behaviors, prioritizing safety or efficiency depending on the current driving scene. 
\begin{equation}
    R_{final} = g_{safe} + g_{efficiency} + g_{smooth}
    \label{eqn:final_reward}
\end{equation}
The following section will explain the reward labeling process, and this includes how the adaptive safety component $c_t$ is determined, as well as how the final reward signal is calculated. 

\section{Reward Labeling Method}
\label{section:reward_labeling}
\begin{figure}[!th]
\centering
\includegraphics[width = 1.0\linewidth]{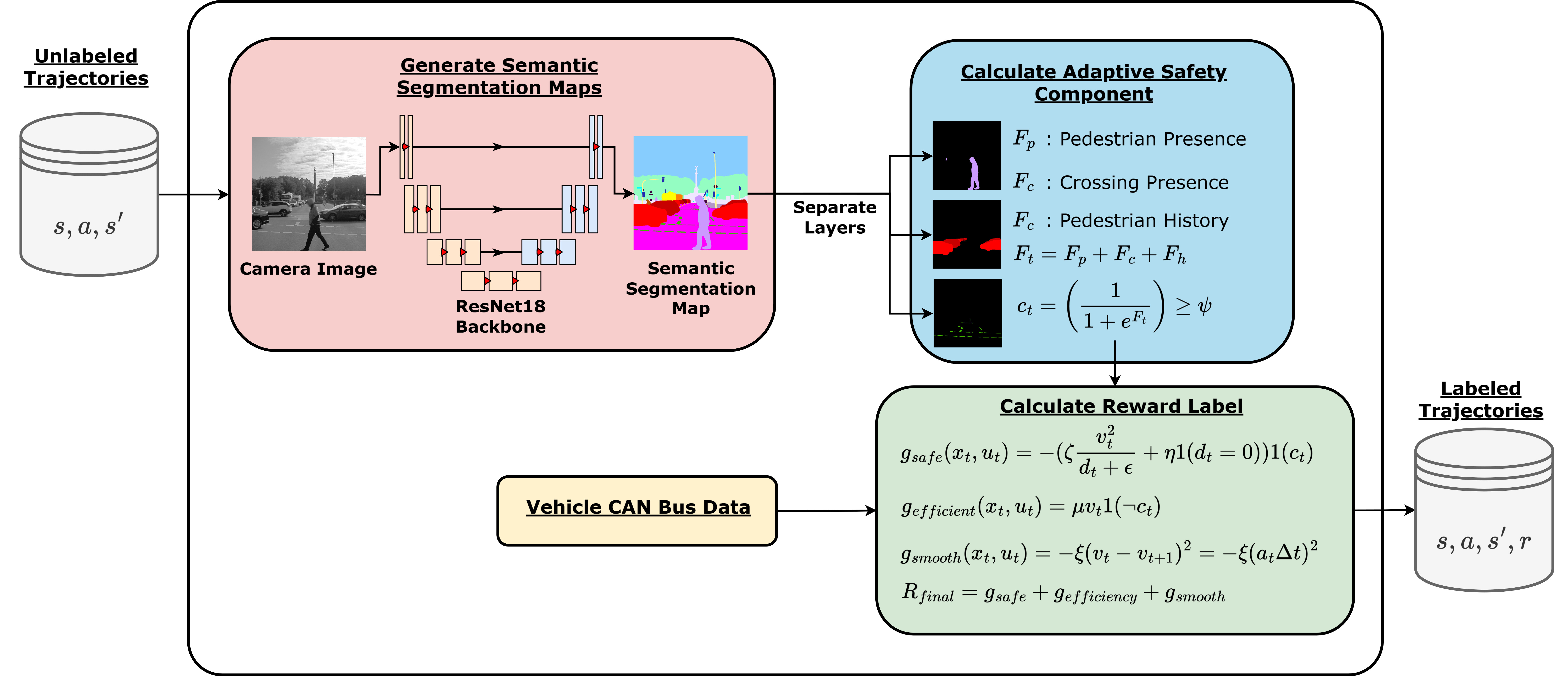}
\caption{An illustration of the proposed reward labeling method for perception-based Offline RL training. Beginning with an unlabeled dataset, camera images are passed through a ResNet18 UNet model to generate semantic segmentation maps. Task relevant layers of the maps like pedestrian, vehicle and zebra crossing are then used to calculate the adaptive safety component $c_t$ of the reward function. The final reward label is calculated with this component and data from different onboard vehicle sensors using Equation \ref{eqn:final_reward}.}
\label{fig:reward_labeling}
\end{figure}

This section introduces a novel method to generate meaningful reward labels that enable existing datasets to be used for Offline RL training. This method focuses on how image data can be leveraged to produce grounded reward labels that reflect the context of the driving scene. To achieve this, the images are processed to generate semantic segmentation maps, that provide pixel-level class labels to distinguish between different entities. Specific class layers from the semantic map that are relevant to task (e.g., pedestrians, zebra crossing, vehicles etc.) can be isolated and analyzed to evaluate the driving scene. The output from this scene evaluation is the adaptive safety component $c_t$, which is used in the reward function (Equations \ref{eqn:safe} and \ref{eqn:efficient}) to switch the appropriate component on. Alongside $c_t$, data from different vehicle sensors are utilized to calculate the final reward labels using the aforementioned reward function. Figure \ref{fig:reward_labeling} provides an overview of the proposed reward labeling method, and the following sub-sections provide a detailed breakdown of each step. 

\subsection{Semantic Segmentation Maps}
\label{section:method_semantic_maps}
One method to efficiently extract information from the driving scene is through semantic segmentation maps. Semantic segmentation involves partitioning an image into regions and providing pixel-level classification for different object classes. Formally, given an input image $\mathbf{I} \in \mathbb{R}^{H \times W \times C}$, the goal is to produce an output segmentation map $\mathbf{S} \in \mathbb{R}^{H\times W}$, where each element/pixel $s_{i,j}$ in $\mathbf{S}$ corresponds to a class label from a set of $K$ possible classes.

To generate the semantic segmentation maps, grayscale image observations (dimensions $1 \times 224 \times 224$) are passed to a UNet model \cite{ronneberger2015u} with a ResNet18 \cite{he2016deep} backbone, employing an encoder-decoder architecture with skip connections. The encoder path of the UNet model captures context through convolutional and max-pooling layers, reducing the spatial dimension while increasing the feature depth. 
The UNet's decoder path reconstructs the image using upconvolutions and combines high-resolution features from the encoder to ensure precise localization, resulting in the final semantic map, which captures $28$ different classes. The $28$ class labels include key objects in the driving scene like pedestrians, zebra crossings, vehicles, and lane markings. The output of the UNet model is a multi-dimensional image array (dimensions $28 \times 224 \times 224$) and each pixel indicates the probability of that pixel belonging to a class. The motivation behind using the UNet model is its ability to produce semantic maps in real-time. This is of utmost importance, as Section \ref{section:method_spatial_attention} will explain how the semantic maps are utilized in the feature extraction stage during Offline RL training. It should be noted that, for the sole purpose of reward labeling, a larger model can be used to produce the semantic maps, as real-time generation is less essential. The key benefits of producing semantic maps are that they offer a greater scene understanding and easy access to the location of different objects in view, as well as their relationships. The next sub-section provides key details on how the semantic maps are analyzed to produce the adaptive safety component $c_t$. 

\subsection{Adaptive Safety Component}
\label{section:method_adaptive_safety_component}
\begin{figure}[!h]
\centering
\includegraphics[width = 0.6\linewidth]{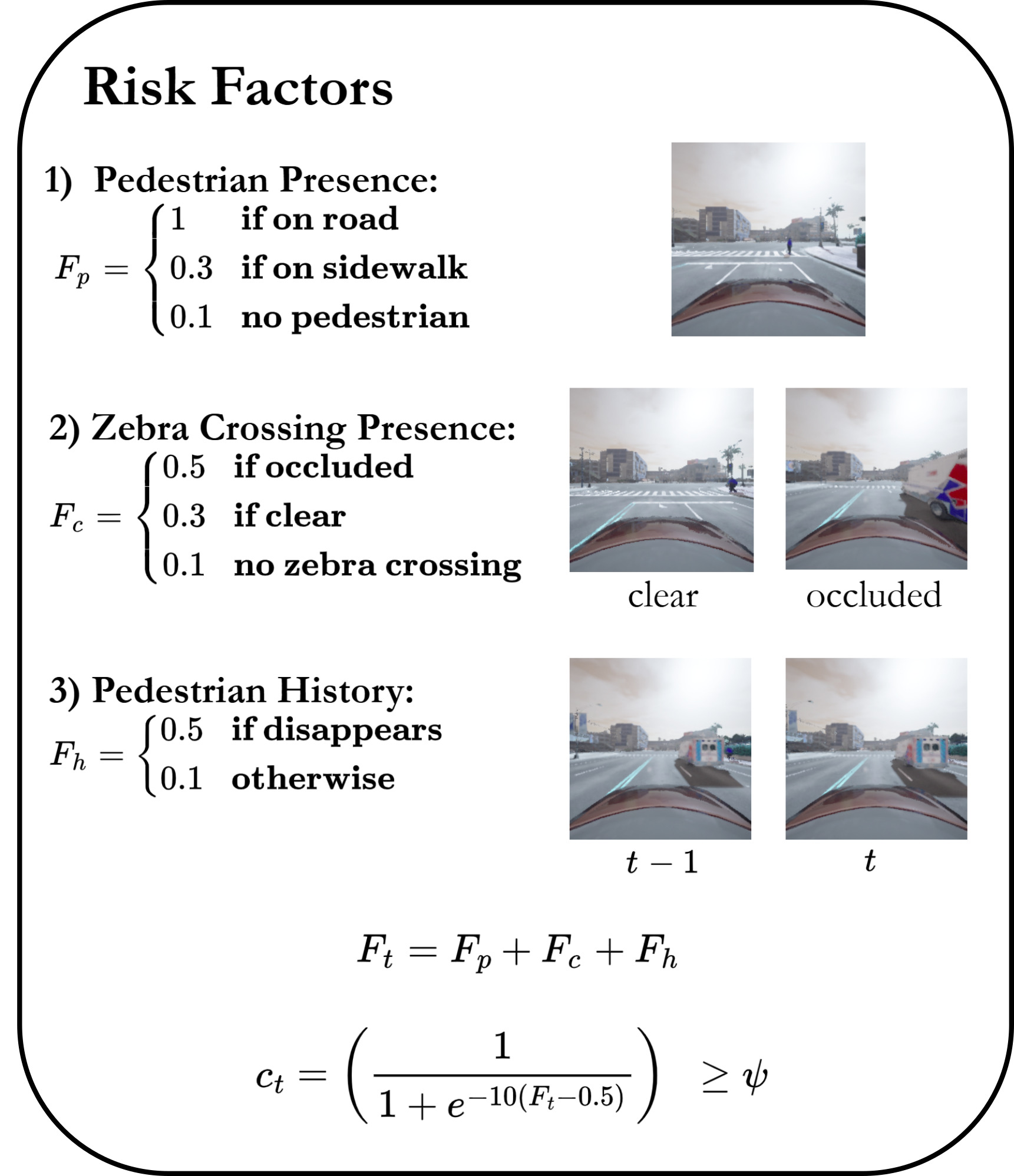}
\caption{An overview of the $3$ risk factors used to determine the adaptive safety component $c_t$, along with camera images of examples of each occurrence. The $3$ risk factors include: 1) Pedestrian Presence, which will be high if there's a pedestrian on the road and low if not, 2) Zebra Crossing Presence, which will be medium if the zebra crossing is occluded and low if clear, and 3) Pedestrian History, which will be medium if a pedestrian is detected and then suddenly disappears. The total risk is the sum of the risk factors, and this value is passed through a scaled sigmoid function. The resultant value is between the range $[0,1]$ and if above the threshold $\psi=0.75$, then $c_t = 1$, otherwise $c_t = 0$.}

\label{fig:risk_factors}
\end{figure}

The adaptive safety component $c_t$ is derived from three pre-defined risk factors, which are characteristics of the driving scene that require the agent to exercise caution. The greater the number of risk factors present, the greater the probability that the agent should prioritize safety over efficiency. Once the probability passes a certain threshold, the adaptive component is set to true ($c_t=1$), causing the safety reward component (Equation \ref{eqn:safe}) to switch on and the efficiency component (Equation \ref{eqn:efficient}) to switch off. Figure \ref{fig:risk_factors} depicts the three risk factors that determine $c_t$, alongside example camera images of each occurrence. The rest of this sub-section explains how the semantic maps (outlined in Section \ref{section:method_semantic_maps}) are used to extract each risk factor.

The first risk factor ($F_p$) considers pedestrian presence in the driving scene; if there's a pedestrian on the road, they pose a high risk and warrant slowing down to avoid collision. However, if the pedestrian is on the sidewalk, then the risk is lower to prevent the agent from driving too cautiously. The method analyzes the semantic map's pedestrian, road, and sidewalk classes to identify the pedestrian's position relative to any road or sidewalk. After detecting a pedestrian in the semantic map, the system analyzes the $10$ neighboring pixels along the border of each pedestrian detection. It counts the number of pixels belonging to the road or sidewalk class and the class with the greater number of pixels is assumed to be the surface the pedestrian is on. If multiple pedestrians are detected on both the sidewalk and the road, the system will register the highest risk factor, as that information is most important to the decision-making process.

The next two risk factors cover more subtle indicators that human drivers may use to anticipate risky situations in busy traffic scenes. The second risk factor ($F_c$) focuses on the presence of a zebra crossing in the agent's view; if the zebra crossing is partially occluded through traffic congestion, then there will be a medium risk, as the agent is unaware of any potential obstacles. Consequently, if the zebra crossing is in full view, then the agent can adequately react to any potential obstacles and thus is assigned a low risk. To achieve this, the method analyzes the semantic map's zebra crossing and vehicle classes. If a zebra crossing is detected, the system gauges scene congestion by examining the vehicle class. A significant number of vehicle pixels results in the zebra crossing being labeled as occluded.

The third risk factor ($F_h$) flags instances where a pedestrian is detected in the driving scene at $t-1$, but disappears in the consecutive timestep, which is treated as a medium risk to alert the system that the pedestrian is occluded and could potentially reemerge. This is achieved through the use of a $10$ cell memory bank, $\mathcal{M} = [m_0, \ldots, m_9]$ and each cell, $m_i$ can hold a binary value: $m_i \in {0, 1}$, where $1$ denotes the presence of a pedestrian and $0$ denotes no pedestrian present in the scene. At each timestep $t$, the current visibility of the pedestrian is recorded in the first memory cell, $m_0$, and the previous values are shifted to the next cells: $m_{i}(t) = m_{i-1}(t-1)$ for $i= 1,2,\ldots, 9$. If no pedestrian was detected in cell $m_{0}(t)$, but a pedestrian was detected in cell $m_{1}(t)$, then a medium risk factor will be registered.

Finally, the total risk, $F_t$ is calculated as the summation of the three risk factors ($F_p$, $F_c$, and $F_h$), and the resultant value is fed into the scaled sigmoid function to output a probability value $\mathcal{P}\in[0,1]$ . To determine whether the situation is high-risk, the probability $\mathcal{P}$ is compared against a predefined threshold $\psi=0.75$ and if $\mathcal{P} > \psi$ then $c_{t} = 1$, otherwise $c_{t} = 0$. The following sub-section explains how the final reward label is calculated using the adaptive safety component and other vehicle sensor data.

\subsection{Calculate Reward Label}
\label{section:method_calculate_reward}
Once the adaptive safety component $c_t$ is calculated, data from onboard vehicle sensors is needed to calculate the reward label using the reward function outlined in Section \ref{section:method_reward_function}. The accelerometer from an Inertial Measurement Unit (IMU) sensor provides the vehicle's acceleration, which is used for the smoothness component (Equation \ref{eqn:smooth}). The vehicle's speed, used in the safety and efficiency reward components (Equations \ref{eqn:safe} and \ref{eqn:efficient}), is typically available on the vehicle CAN bus in most existing AD datasets.

Furthermore, the safety component (Equation \ref{eqn:safe}) uses the distance between the detected pedestrian and the vehicle to penalize the agent for traveling at a high speed towards the pedestrian. In the case of simulation-based setups, the distance information can be directly obtained from the simulator; however, for real-world data, the distance between the pedestrian and the vehicle must be inferred from sensor data. This information can be obtained via Radar sensors with high accuracy, however, it is difficult to distinguish between different objects in view. Alternatively, inferring this distance from camera images using the pinhole method in computer vision enables the system to infer the distance of specific objects in the scene.
The pinhole method is a fundamental technique in computer vision used to estimate distances in a scene. It models the camera as a pinhole, where light rays pass through a small aperture and project an inverted image on the opposite side. By using the camera's intrinsic parameters, such as focal length and sensor size, and the known size of objects in the scene, the distance $D$ to an object (in this case, a pedestrian) can be estimated. The relationship between the actual height of the pedestrian $H$, the height of the pedestrian in the image $h$, and the distance $D$ can be expressed as: 

\begin{equation}
    D = \frac{H \cdot f}{h}
\end{equation}

\noindent where $f$ is the focal length of the camera. By analyzing the dimensions and positions of pedestrians detected in the camera's images, the pinhole method enables the estimation of their distances from the vehicle in real-time.

Altogether, the data from the aforementioned vehicle sensors and the adaptive safety components $c_t$ are used to calculate the reward labels for an unlabeled dataset of driving trajectories using the reward function defined in Equation \ref{eqn:final_reward}. The following section will provide further details on the Offline RL training procedure and how the semantic segmentation maps are used to provide varying levels of spatial attention for different classes of objects.

\section{Offline Reinforcement Learning Training}
\label{section:offline_rl}
\begin{figure}[!h]
\centering
\includegraphics[width = 1.0\linewidth]{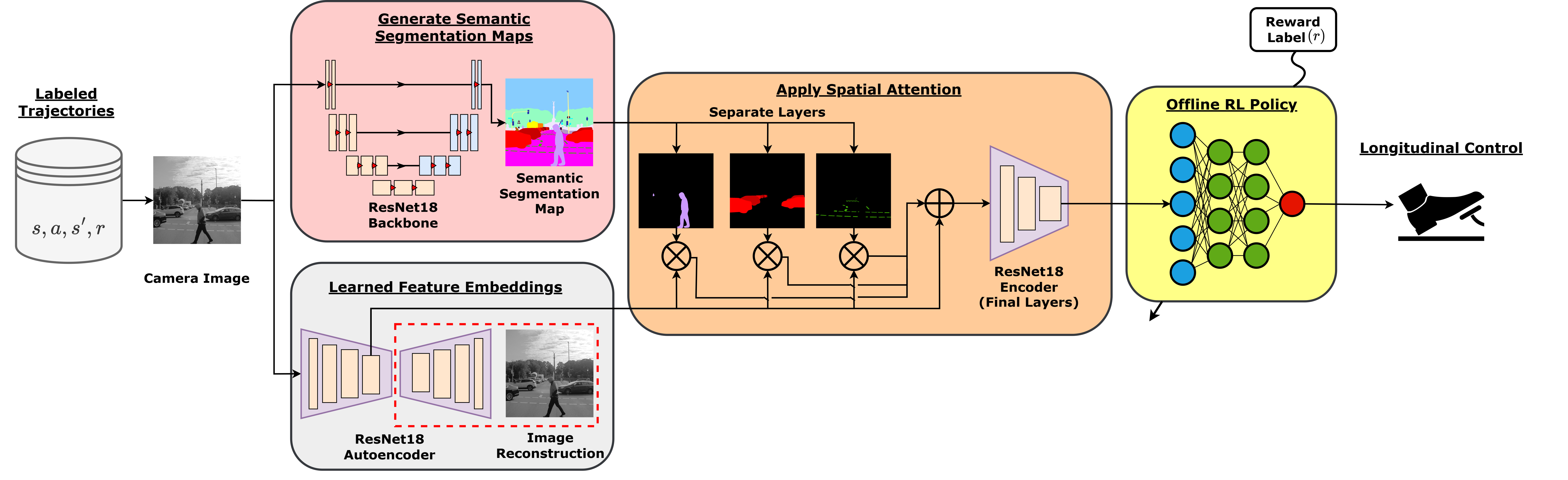}
\caption{An illustration of the proposed Offline RL training process to highlight different objects in the driving scene to varying degrees. The pipeline begins with passing camera images from the labeled dataset to a ResNet18 autoencoder (decoder model highlighted in red is removed during inference) to obtain the latent feature embeddings. In addition the camera images are passed through the same ResNet18 UNet model outlined in Section \ref{section:method_semantic_maps} to generate the semantic segmentation maps. Next the task relevant semantic layers are used to provide spatial attention to the latent feature embeddings, which are then passed downstream to train the Offline RL policy. The final output is the vehicle longitudinal control.}
\label{fig:offline_training}
\end{figure}

This section formally introduces a novel method to improve the training efficiency of end-to-end perception based systems via utilizing the class layers of semantic segmentation maps to highlight key objects in the latent feature embeddings. In RL settings with low-dimensional state spaces, agents can efficiently learn a policy, as the state representation already captures key information needed to complete the task. In contrast, high dimensional image-based states often contain large amounts of redundant or irrelevant information, which leads to longer training times, as the agent must infer what features are important. By highlighting the salient objects within the image data, the agent can be guided to focus on the most relevant information for decision-making. This approach is inspired by how humans naturally focus on visual cues like other vehicles, zebra crossings and pedestrians when driving. Figure \ref{fig:offline_training} illustrates the proposed pipeline, which includes the generation of semantic segmentation maps outlined in Section \ref{section:method_semantic_maps}. The following sub-section will explain the feature extraction process, and includes the method of applying spatial attention.

\label{section:method_offline_rl}

\subsection{Spatial Attention using Semantic Segmentation Maps}
\label{section:method_spatial_attention}
The proposed Offline RL training pipeline extends previous works \cite{asodia2024human} and is illustrated in Figure \ref{fig:offline_training}. Following the reward labeling process outlined in Section \ref{section:reward_labeling}, camera images from the labeled trajectories are passed through two streams. The first stream generates the semantic segmentation maps in the same manner outlined in Section \ref{section:method_semantic_maps}. The second stream consists of a ResNet18 autoencoder, where the images are passed through an encoder, producing latent feature embeddings. The purpose of the encoder is to produce a compact representation of the edges, shapes, and objects within the original image. These feature embeddings are then passed through a decoder model to reconstruct the original image, where the quality of the reconstructed image indicates the accuracy of the feature embeddings. During inference, the decoder model is no longer needed and is discarded. The motivation behind using ResNet18 as the backbone for the autoencoder, as well as the UNet model, is that it is relatively lightweight, using fewer layers, which corresponds to faster inference times.

The next stage applies spatial attention to the latent features using different layers from the semantic segmentation maps. Specifically, the pedestrian, zebra crossing, and vehicle semantic layers $Z_{ped}, Z_{cross}, Z_{veh} \in \mathbb{R}^{1 \times 224 \times224}$ are isolated. The justification for selecting these layers is that human drivers typically focus on these objects when navigating complex driving scenarios like occluded pedestrian crossings. Thus, it stands to reason that these objects' features should be highlighted to enable the downstream Offline RL agent to efficiently learn the task of vehicle longitudinal control. 

The spatial attention is applied over a two-step process; the first step consists of a weighted cross-wise multiplication between the different semantic layers (resized to $1\times56\times56$ using bilinear interpolation) and the feature cube $Z_c \in \mathbb{R}^{256 \times 56 \times56}$ from the penultimate layer of the ResNet18 encoder. Each cross-wise multiplication operation is weighted by the relative importance that the semantic class has for the task of collision avoidance. This weighting process is motivated by how humans would typically focus on specific objects on the road when driving. For example, the highest weighting is assigned to the pedestrian class ($\omega_{ped}=1.0$); as the most vulnerable actors on the road, the typical driver would pay close attention to them to anticipate their movement. A similar rationale can be applied to the zebra crossing class: during times of heavy congestion, drivers must be vigilant for any crossing pedestrian, which is reflected in a weighting of $\omega_{cross}=0.75$. Finally, the lowest weighting, $\omega_{veh}=0.50$, is applied to the vehicle class, as the agent should acknowledge surrounding vehicles; however, compared to the aforementioned classes, their features are less integral to the decision-making process. The outcome from each cross-wise multiplication operation (Equations \ref{eqn:spatial_attention_step_1_a}-\ref{eqn:spatial_attention_step_1_c}) is a set of intermediate features $\hat{Z}_{ped}, \hat{Z}_{cross}, \hat{Z}_{veh} \in \mathbb{R}^{256 \times 56 \times56}$ that are used in the second step.
\begin{subequations}
\begin{align}
\hat{Z}_{ped}   &= \omega_{ped} \cdot (Z_{ped} \otimes Z_c) \label{eqn:spatial_attention_step_1_a} \\
\hat{Z}_{cross} &= \omega_{cross} \cdot (Z_{cross} \otimes Z_c) \label{eqn:spatial_attention_step_1_b} \\
\hat{Z}_{veh}   &= \omega_{veh} \cdot (Z_{veh} \otimes Z_c) \label{eqn:spatial_attention_step_1_c}
\end{align}
\label{eqn:spatial_attention_step_1}
\end{subequations}
The final step to apply the spatial attention is to perform a cross-wise addition operation between the set of intermediate features $\hat{Z}_{ped}, \hat{Z}_{cross}, \hat{Z}_{veh}$ and the original feature cube $Z_c$ (Equation \ref{eqn:spatial_attention_step_2}). This results in a final feature cube $Z_{spatial} \in \mathbb{R}^{256 \times 56 \times56}$ that has different salient regions boosted, relative to their importance for the given task. This feature cube is then sent through the remaining layers of the ResNet18 encoder to produce a compact 1D array representation, which is then passed downstream to train the Offline RL policy. The final output of the pipeline is the vehicle longitudinal control action.
\begin{equation}
    Z_{spatial}= \hat{Z}_{ped} \oplus \hat{Z}_{cross} \oplus \hat{Z}_{veh} \oplus Z_c
    \label{eqn:spatial_attention_step_2}
\end{equation}

\section{Experimental Setup}
\label{section:experimental_setup}
This section outlines the experimental setup used to evaluate both the reward labeling method and the proposed Offline RL training pipeline. Key details are provided on the driving scenario used in this study, the comparative baseline reward labeling methods, and the training process. The first set of experiments aims to evaluate the policies trained on the labeled dataset in the CARLA Urban Driving simulated environment. CARLA provides a wide set of dynamic actors and sensors, enabling practical training and evaluation of AVs in urban scenarios. Evaluating offline policies often involves online interactions to assess their decision-making more accurately. However, fully offline evaluation remains challenging due to distributional shifts, making it difficult to reliably estimate the value of an action in an unseen state without additional interactions with the environment. The second set of experiments moved away from the simulated environment and involved using the reward labeling method on the A2D2 dataset. The reward labels were compared with those produced by a human annotator, primarily to evaluate the proposed method’s decision-making when applied to real-world data. This section will provide details on both the subset of real-world data used in the experiment and the human annotation process.

\subsection{Occluded Pedestrian Crossing Scenario using CARLA}
\label{section:scenario}

\begin{figure}[!th]
      \centering
      \setlength{\fboxsep}{0pt}%
      \setlength{\fboxrule}{0pt}%
      \framebox{\parbox{3in}{\includegraphics[width=\linewidth]{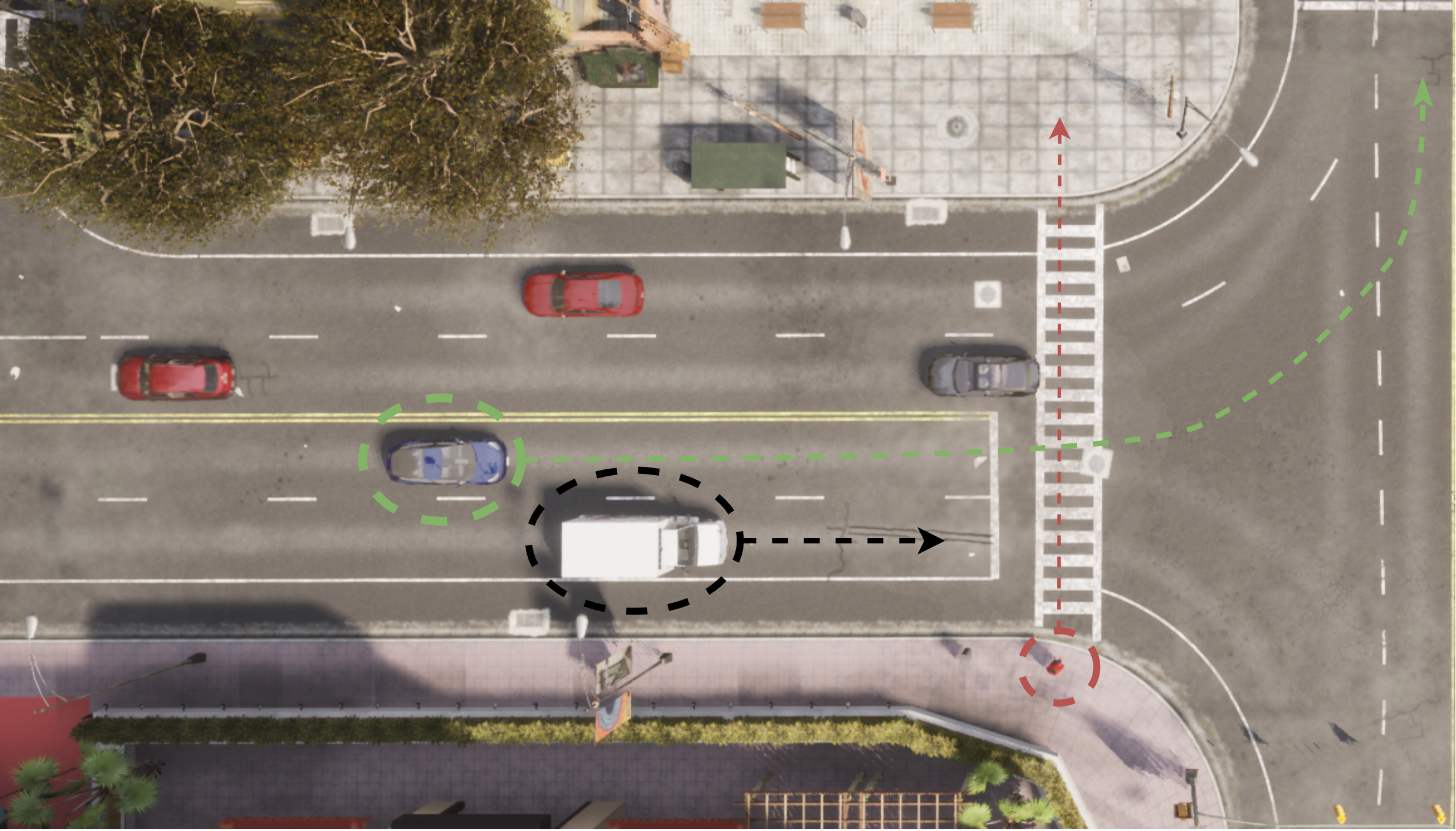}}

}
      \caption{Occluded pedestrian crossing scenario setup within CARLA. The ego vehicle, highlighted in green, is spawned at one end of the road and must navigate through a zebra crossing occluded by a vehicle, highlighted in black. The ego vehicle must yield for a crossing pedestrian, highlighted in red, to successfully reach the goal point at the other end of the road.}
      \label{fig:carla-eb}
   \end{figure}

This study used the occluded pedestrian crossing scenario to evaluate the proposed work, as the scenario reflects a high-risk, safety-critical situation in which AVs must anticipate hidden hazards and respond conservatively. Figure \ref{fig:carla-eb} illustrates the occluded pedestrian crossing scenario setup in CARLA. In this scenario, the ego vehicle starts at one end of the road and must reach the goal at the other end. There are two variations: in one, a pedestrian is initially visible but becomes occluded by a van before suddenly crossing the road as the vehicle approaches the zebra crossing; in the other, the pedestrian remains occluded until they suddenly cross. The goal of the RL agent is to adjust the vehicle's longitudinal speed to safely and efficiently reach the goal point while avoiding collisions with the pedestrian.

As a stress test, the pipeline is evaluated with varying pedestrian traffic: `Low Ped Traffic' with $10$ pedestrians, `Med Ped Traffic' with $25$ pedestrians ($20\%$ running at $6km/h$ and $20\%$ crossing unexpectedly), and `High Ped Traffic' with $50$ pedestrians ($40\%$ running at $6km/h$ and $30\%$ crossing unexpectedly). The motivation for evaluating the pipeline across different congestion levels is to gauge the effectiveness of the risk factors used to determine the adaptive safety component $c_t$ (outlined in Section \ref{section:method_adaptive_safety_component}).

The ego vehicle was equipped with an RGB dashboard camera for image observations, a semantic dashboard camera for data collection, and an IMU sensor for kinematic data. Moreover, the ego vehicle has been fitted with a custom sensor available in CARLA, designed to detect collisions between the vehicle and other objects, such as pedestrians and other road users.

\subsection{Dataset}
\label{section:dataset}
To train the ResNet18 UNet model, $20k$ images and $20k$ semantic labels were collected from the RGB and semantic cameras in CARLA. The original semantic camera in CARLA uses $27$ classes to separate objects in the scene. However, in the original setup, zebra crossings were part of the road marking class, and for the risk factors outlined in Section \ref{section:method_adaptive_safety_component}, only the zebra crossing road markings were considered. Thus, the zebra crossings were separated into their own class, making $28$ classes in total. To ensure that the UNet model generalizes well, the ego vehicle's and pedestrian's positions, including longitude, latitude, altitude, and the vehicle's rotation, were uniformly randomized for each sample.

To train the Offline RL policies for the simulation-based experiments, transitions over $100$ episodes were collected in CARLA and processed by the proposed reward labeling method (illustrated in Figure \ref{fig:reward_labeling}), resulting in a final dataset of $22101$ samples, $\mathcal{D}=\{(s_i,a_i,s_i',r_i)\}^{\mathcal{N}=22101}_{i=1}$. Given that the scope of the experimentation is limited to the localized pedestrian crossing scenario, a dataset of this size is sufficient for learning an optimal policy.

\subsection{Model Evaluation} 
\label{section:model_evaluation}
This sub-section presents the different reward labeling methods and Offline RL algorithms used in the simulation-based experiments. The purpose of the different reward labeling methods is to serve as comparative baselines to the proposed method, and the purpose of using different Offline RL algorithms is to evaluate each reward labeling’s effectiveness across multiple algorithms. It will provide details on the collision tests that quantitatively assess the policies, as well as information on the trajectory analysis, which offers a qualitative assessment of the policies.
The reward-labeling methods are as follows:
\begin{itemize}
    \item \textbf{Gen:} This is the proposed reward labeling method.
    \item \textbf{Unlabeled Data Sharing (UDS) \cite{yu2022leverage}:} The authors proposed that a small dataset with meaningful reward labels can be supplemented by assigning zero rewards to additional unlabeled data, allowing the combined dataset to support effective policy learning. This baseline was implemented by randomly assigning zero rewards to $50$ episodes of the original CARLA dataset, with the remaining episodes being labeled with rewards in CARLA.
    \item \textbf{Vision Language Model (VLM):} This baseline combines the VLM, VideoLLaMA3-2B-Image, with Retrieval-Augmented Generation (RAG). It generates reward labels by comparing the current state with labeled reference trajectories and inferring a scalar reward consistent with the $3$ most similar reference states. The reference trajectories had been collected using a mixture of policies: behavioral, random, aggressive, and conservative. The reward labels for these reference trajectories had been taken from CARLA. At each time step $t$, a sequence of image frames centered at $t$ is fed into the VLM. Along with these frames, the constructed textual prompt shown in Figure \ref{fig:vlm_prompt} is provided. This prompt aggregates current kinematic information and the retrieved nearest-neighbor precedents. The VLM then processes this multimodal input to generate a natural-language response. This response is subsequently parsed to extract a scalar reward $r \in [-1, 1]$.
\end{itemize}

\begin{figure}[!bh]
\centering
\fbox{
\begin{minipage}{0.95\linewidth}
\small\ttfamily
You are an expert autonomous driving evaluator.\\[4pt]

Current vehicle state:\\
\textless STATE\_ITEMS\textgreater\\[4pt]

Nearest neighbor states and their rewards:\\
1. \textless NEIGHBOR\_STATE\_1\textgreater\\
2. \textless NEIGHBOR\_STATE\_2\textgreater\\
3. \textless NEIGHBOR\_STATE\_3\textgreater\\

You are shown one or more consecutive frames from the vehicle's front camera.
Considering safety, comfort, and efficiency, predict the REWARD for the current state.\\[4pt]

Respond strictly in the following format:\\
REWARD: \textless float in [-1, 1]\textgreater\\
REASONING: \textless 1--2 concise sentences\textgreater
\end{minipage}
}
\caption{Prompt template used for the VLM reward labeling baseline.}
\label{fig:vlm_prompt}
\end{figure}

To gauge how the aforementioned reward labeling methods perform across different Offline RL algorithms, the following methods were used in the evaluation process:
\begin{itemize}
    \item \textbf{Conservative Q-Learning (CQL) \cite{kumar2020conservative}:} It utilizes a conservative penalty in the critic objective to reduce Q-values for OOD actions. The pessimistic value estimates limit over-optimistic policy updates, providing greater performance stability.
    \item \textbf{Implicit Q-Learning (IQL) \cite{kostrikov2021offline}:} It avoids explicit behavior constraints by using expectile regression to estimate the state value function. Policy improvement then follows advantage-weighted regression, which prioritizes high-value actions in the dataset without querying OOD actions.
    \item \textbf{Behavioral Proximal Policy Optimization (BPPO) \cite{zhuang2023behavior}:} It adapts the Proximal Policy Optimization algorithm to an offline setting, via monotonic improvements to a behavioral policy $\pi_b$. The algorithm restricts policy updates that deviate significantly from the behavioral policy.
\end{itemize}

Together, the simulation-based experiments include $9$ setups, comprising different permutations of reward labeling methods and Offline RL algorithms. To note, each setup will incorporate the Offline RL pipeline detailed in Section \ref{section:offline_rl}. Going forward, to reference a setup, the following notation will be used: "IQL (Gen)", which corresponds to an IQL model trained with rewards generated with the proposed method. 

The simulation-based experiments begin with an analysis of the rewards produced by the different methods, aiming to contrast and compare their distributions. Due to the high dimensionality of the observation space, the Uniform Manifold Approximation and Projection (UMAP) technique is used for dimensionality reduction, mapping each image to a single scalar value. Similar image observations will have UMAP representations that are close proximity to each other.

For the quantitative evaluation, the training curves for each setup were captured. Collision tests are also conducted for each setup in the occluded pedestrian crossing scenario, with the success, collision, and timeout rates recorded, along with the average stopping distance over $100$ evaluation episodes. The collision tests are conducted under varying levels of pedestrian traffic, as detailed in Section \ref{section:scenario}, and are adjusted for different levels of occlusions. In the fully occluded setting, a large van was placed near the zebra crossing, while a smaller vehicle was used in the partial occluded setting. These collision tests aim to gauge the system's performance in complex, unpredictable traffic scenes and to assess how different levels of occlusion affect its decision-making.

On the other hand, for the qualitative evaluation, a trajectory from the high pedestrian traffic scenario was collected, and the risk factors used to determine the adaptive safety component $c_t$ were logged to visualize their impact on the reward along the trajectory.

\subsection{Training Details} 
This sub-section provides the training details that facilitated the simulation experiments.  
To train the ResNet18 UNet model for semantic segmentation, a weighted sum of the Cross-Entropy Loss and Dice Loss has been used, enhancing both pixel-wise classification accuracy and the overall segment structure. The Adam optimizer has been used along with a step learning rate scheduler, which reduced the initial learning rate of $0.001$ by a factor of $10$ every third of the training duration. The ResNet18 UNet model has been trained for $45$ epochs, which proved sufficient, as the performance on the downstream vehicle longitudinal control task showed no meaningful gains with additional training. 

All training and processing have been performed on an NVIDIA GeForce RTX 4060ti graphics card. On the whole, the training times for the Gen and UDS reward setups were very similar, with the CQL models taking $\approx8$ hours, the IQL models taking $\approx11.5$ hours, and the BPPO models taking $\approx14.5$ hours to train. It should be noted that the training times for the VLM reward setups took considerably longer. Both the Gen and UDS methods took $0.5$ hours to label the collected data, whereas the VLM method with the VideoLLaMA-2b model took roughly $60$ hours. The training times for the ResNet18 encoder-decoder and UNet model were $6$ and $1$ hours, respectively.

\subsection{Real World Application}
\label{section:real-world-application}
The first set of simulation-based experiments modeled the occluded crossing scenario under varying levels of pedestrian traffic. However, these settings do not fully capture the wide range of conditions encountered in real-world driving; thus, the method should be applied to real-world AD data to provide a more comprehensive picture of its capabilities. Moreover, the previous test set did not assess the method's decision-making relative to human judgment. To address this, additional tests were carried out, applying the reward labeling method to a subset of real-world AD data and comparing the labels to those produced by a human annotator. This sub-section will provide key details on the subset of data used and a definition of the human annotator.

The chosen dataset was the A2D2 open-source dataset, which contains multi-sensor data collected from real-world urban driving scenarios. The dataset comprises high-resolution images from multiple cameras, 3D point clouds from LiDAR, GPS data, and annotated semantic segmentation labels for objects such as vehicles, pedestrians, and road markings. In addition, the A2D2 dataset contains sensor data from a built-in car gateway, including the vehicle's speed and acceleration at each timestep, providing sufficient information to generate reward labels with the method. The reward labeling method was applied to a small subset ($1315$ samples) of the A2D2 dataset, in which trajectories that included pedestrians crossing were isolated. Creating this subset was done because the study's scope is focused on the crossing scenario.

The main aim of this experiment was to compare the proposed risk-factor-based approach (detailed in Section \ref{section:method_adaptive_safety_component}) with human decision-making in determining unsafe observations. To do so, a human annotator served as a baseline for comparison. In this study, a "human annotator" is defined as a human who manually annotated a subset of the A2D2 data, identifying risky situations. Specifically, the human annotator was tasked with manually labeling the adaptive safety component, $c_t$, for each sample. They assigned $c_{t}$ the value $1$ if they determined that the vehicle should yield to the pedestrian. This included samples where pedestrians were either about to cross or in the process of crossing the road, as well as situations where the annotator believed caution was necessary to avoid potential collisions. In contrast, samples showing normal driving conditions with no risk of collision were labeled as safe ($c_t=0$). After manually determining the adaptive safety components, the proposed reward function (Equation \ref{eqn:final_reward}) was used to calculate the final reward. 

\section{Results}
\label{section:results}
This section presents results from simulation-based and real-world experiments, focusing on the effectiveness of reward labels for Offline RL training.

\subsection{Reward Distribution Analysis}
\begin{figure}[t] 
    \centering 
    \subfloat[State-reward distribution plot.\label{fig:state-reward-dist}]{ \includegraphics[width=1.0\linewidth]{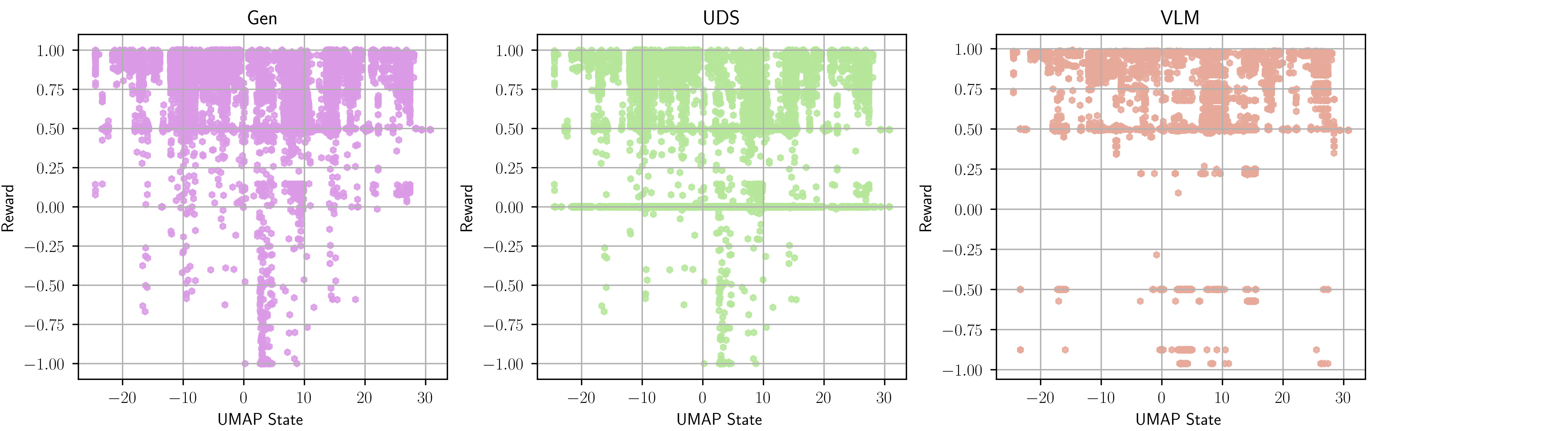} }\\ 
    \subfloat[State-action distribution plot with a color-bar representing the reward value.\label{fig:state-action-reward-dist}]{ \includegraphics[width=1.0\linewidth]{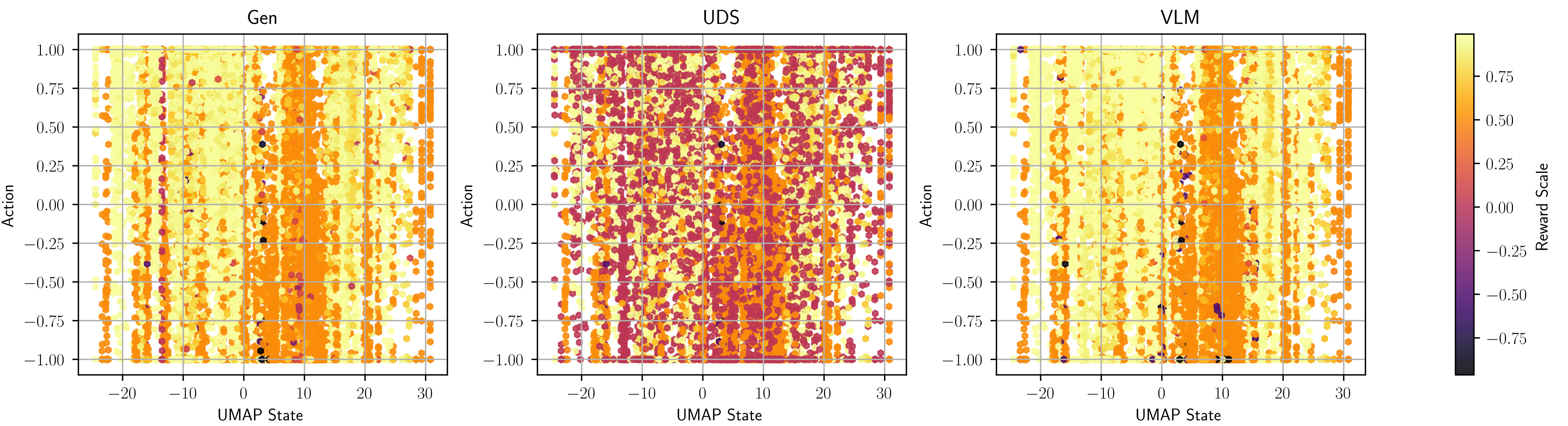} } 
    \caption{A collection of plots showing the distribution of reward labels from each method for the dataset collected in CARLA. The top row of plots shows how the rewards are distributed against the UMAP state representation. The bottom row of plots shows the state-action distribution, with each sample color-weighted by its reward.} 
    \label{fig:distribution-plot} 
\end{figure}

Figure \ref{fig:distribution-plot} illustrates the distribution of reward labels produced by the different methods for the dataset collected in CARLA. The top row of plots (Figure \ref{fig:state-reward-dist}) displays the state-reward distributions. The rewards generated by the proposed method were positively skewed across the UMAP state space. A large portion of states were labeled with rewards $\geq0.50$, which aligns with the dataset's structure: most frames represent instances in which the ego vehicle is not interacting with the zebra crossing. The UDS baseline exhibited the expected behavior, with a dense horizontal band at reward $=0$, as expected given the algorithm's ideology. Interestingly, compared to the proposed method, the VLM rewards exhibit a more distinct clustering of positive rewards, alongside a smaller, separate cluster of negative rewards, without any gradient connecting the two. Among the negative VLM rewards, there is less variation in magnitude, with horizontal bands around approximately $-0.50$, $ -0.85$, and $-0.95$. This results trend implies that the VLM method tends to categorize similar dangerous situations rather than produce a smoother, continuous reward spectrum.

Moving on to Figure \ref{fig:state-action-reward-dist}, by introducing the third variable of vehicle longitudinal control action, the rewards can be visualized across the states and actions. The proposed method exhibits a predominantly positive-valued distribution across most state-action combinations. Visually, there are multiple vertical bands of samples that have rewards of similar magnitude. Although it is difficult to infer which UMAP states correspond to which images, the brighter point clusters in the mid-range of the UMAP state axis indicate that actions taken in common driving states were assigned a moderate-to-high reward value. Additionally, it can be inferred that state-action samples between UMAP states $0$ and $15$ correspond to instances where the ego vehicle is either approaching or at the zebra crossing, as these samples are mostly labeled with rewards between $\approx-1$ and $0.50$. Negative rewards are present, but are relatively sparse and appear scattered across the action range. 

As expected, the UDS reward labels show reward labels of $0$ uniformly distributed across the entire state-action space. Moreover, as shown in Figure \ref{fig:state-reward-dist}, the VLM rewards exhibit a greater contrast between high- and low-valued rewards than the proposed method. There is a denser band of high positive rewards between UMAP states $-17$ and $0$, followed by a distribution of mixed lower rewards similar to that of the proposed method between UMAP states $1$ to $15$. These findings reinforce prior conclusions about the VLM rewards: the method can reliably label distinctively safe and unsafe samples, but may fall short in more complex cases. 

The following sub-section presents the training curves for each Offline RL algorithm trained on the dataset labeled using the different reward labeling methods.

\subsection{Training Performance}
\begin{figure}[!th]
\centering
\includegraphics[width = 0.9 \linewidth]{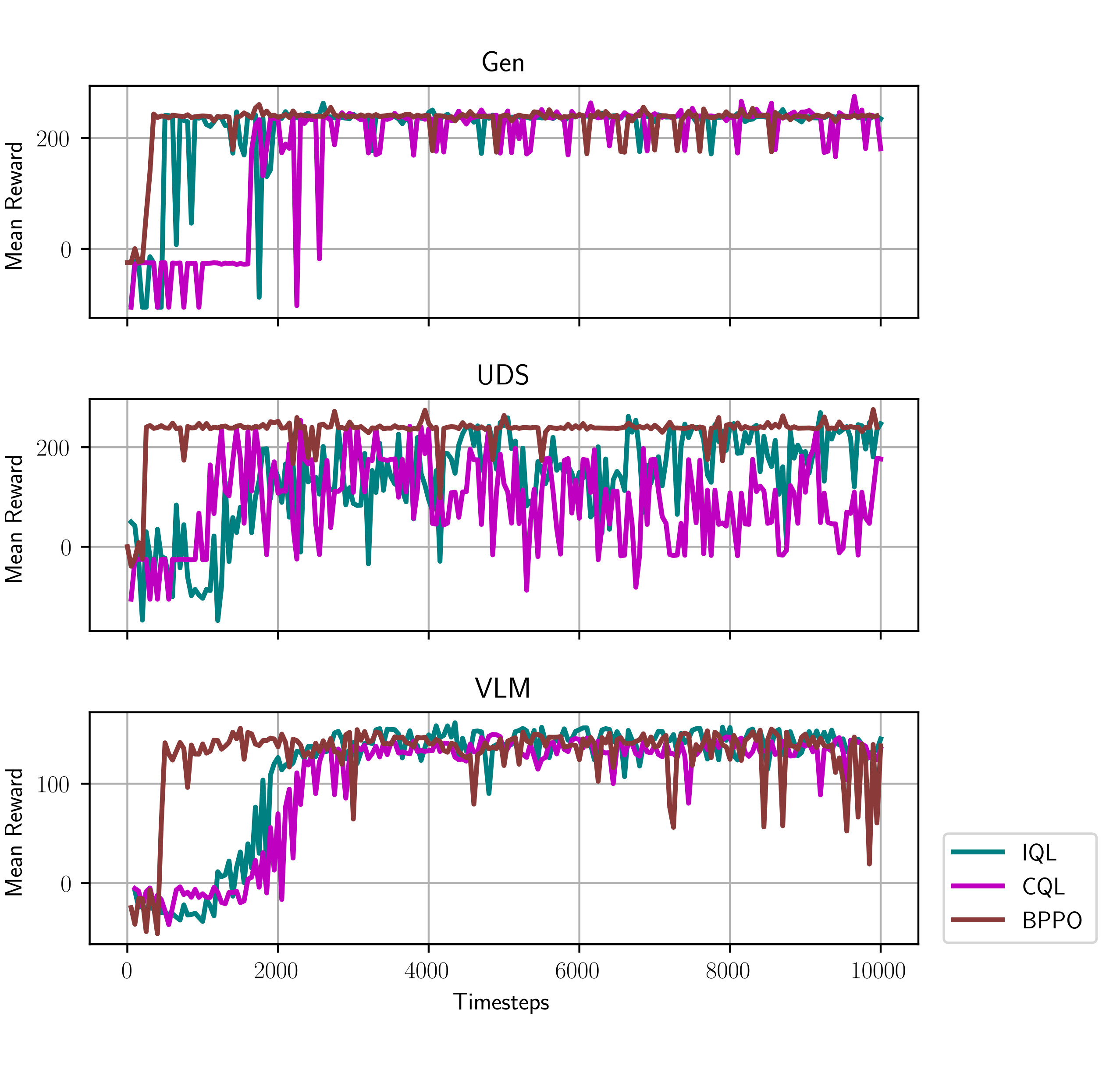}
\caption{Graphs depicting the mean training reward for the different reward labeling methods. For each method, the training results for each Offline RL algorithm are shown, with IQL results in Teal, CQL results in Magenta, and BPPO results in Brown.}
\label{fig:training_reward}
\end{figure}

Figure \ref{fig:training_reward} displays the training performance for the different Offline RL algorithms trained using the different reward labeling methods.

For the Gen Reward setups, the BPPO model showed the fastest convergence and exhibited high stability throughout training. This result is expected, since the algorithm builds on an existing BC policy, whereas IQL and CQL train from scratch. Furthermore, BPPO incorporates the advantage function within training, relying on relative, trajectory-local signals rather than recursive Bellman propagation. The IQL model converged shortly after, leaving the CQL model to be the last to converge. All models maintained their performance, with CQL showing the most dips in mean reward throughout. 

Moving to the UDS setups, the immediate takeaway is that the BPPO model appeared to be least impacted by the $0$ reward labels, again showing fast convergence and stable performance. A reasonable explanation is that the advantage function in BPPO upweights actions with higher-than-average returns, so actions with near-zero returns have less impact during training. On the other hand, algorithms like IQL and CQL that use Bellman backup equations are significantly more affected, as the errors induced by $0$ reward labels can compound over time. This is reflected in the two models’ training curves, which are significantly noisier and show no signs of stable convergence. However, the results show that a learning signal can be extracted, and it should be noted that training stability does not guarantee better performance. Later experiments will provide more clarity on the quality of the learned policies.

Finally, for the VLM setups, the overall trends in the training curves are similar to the Gen Reward setups, but with fewer spikes in the earlier portion of training. Moreover, the scale of the mean training rewards is lower than that of the other two reward labeling methods, as the models converge around a mean reward of $\approx150$. This is primarily due to the VLM assigning different magnitudes of reward per sample, which may have resulted in a different scale of episodic reward. For this reason, it cannot be asserted that the VLM method is worse than the Gen or UDS method.

\subsection{Collision Test}
To evaluate the effectiveness of the generated reward labels, collision tests are conducted. Table \ref{tbl:collision_test} presents the results of the collision test for different Offline RL algorithms (IQL, CQL, and BPPO) trained with rewards generated by the methods (Gen, UDS, and VLM). 

Across all traffic densities, IQL exhibited more consistent high performance than the CQL and BPPO setups. In particular, amongst the IQL setups, IQL (Gen) showed the highest performance under a partial occlusion setting, with success rates ranging from $98\text{-}100\%$ and demonstrated a lesser degradation in performance under fully occluded settings as the traffic density increases. In contrast, the IQL (UDS) setup maintains competitive performance in low-density settings, indicating that the IQL agent was able to leverage unlabeled data for policy training. Similar to the IQL (Gen) setup, a greater decrease in success rates and an increase in stall rates are observed under more congested traffic scenes. This suggests that UDS can be useful in simpler settings; however, unsurprisingly, in more complex scenarios, the agent will benefit from more meaningful reward labels. Moreover, the IQL (VLM) setup showed the weakest performance of the three, with the highest stall rates ($12\text{-}18\%$) under full occlusion, as well as consistently high average stopping distances ($3.95$-$4.66\text{m}$). Whilst the VLM method produced meaningful rewards, it appears that the rewards did not fully capture the intermediate states between explicitly safe and unsafe. 

Overall, the CQL setups showed a substantial performance difference across strategies. Among the three setups, CQL (Gen) achieved the best performance, particularly under partial occlusions, where collision rates are close to $0$ and average stopping distances are relatively long. However, the full occlusion setting exposes a common weakness: success rates drop sharply to $42\%$ at high density, and timeouts become the dominant outcome. Relative to CQL (Gen), CQL (UDS) at low and medium densities tended to underperform; however, at high density, the setup proved more stable under partial occlusions. Interestingly, despite CQL (VLM) showing poor base performance under full occlusions, it exhibited the greatest stability across traffic densities. These results indicate that the proposed method may not generalize to all Offline RL algorithms, but it provides a strong starting point. The pitfalls of the VLM method are its significant computational cost and the black-box nature of reward generation, which can make it challenging to optimize.

The BPPO setups further highlight how different Offline RL algorithms are affected by different reward-labeling methods. Across traffic densities, BPPO (Gen) and BPPO (UDS) performed similarly, demonstrating the viability of the UDS method. In contrast, BPPO (VLM) exhibited the most significant performance degradation across both occlusion levels, as well as the lowest average stopping distance ($1.08\text{-}1.14\text{m}$). These patterns indicate that BPPO tended to learn a more aggressive policy from the given reward labels, and the VLM rewards lead to the most aggressive policies. 

The key takeaway is that reward-labeling quality has a noticeable impact on performance, with Gen reward labels yielding the most stable and reliable behaviors. Among the algorithms, IQL benefited most from reward labels; CQL is highly sensitive to higher occlusion levels; and BPPO notably learns aggressive policies. Across different setups, simple UDS labels yielded sufficient performance in low-complexity settings; however, robust reward generation is essential for generalizable performance in occluded pedestrian scenarios.

\begin{table}[!th]
\centering
\caption{Collision test results from running each setup for $100$ episodes in the occluded pedestrian crossing scenario with varying levels of pedestrian traffic density, where S = Successful, C = Collision, T = Timeout, SD = Stopping Distance, P = Partial Occlusion, and F =  Full Occlusion.}
\label{tbl:collision_test}
\resizebox{\textwidth}{!}{%
\begin{tabular}{c cccccccc cccccccc cccccccc}
\toprule
& 
\multicolumn{8}{c}{\textbf{Low Traffic Density}} &
\multicolumn{8}{c}{\textbf{Medium Traffic Density}} &
\multicolumn{8}{c}{\textbf{High Traffic Density}} \\
\cmidrule(lr){2-9} \cmidrule(lr){10-17} \cmidrule(lr){18-25}
& \multicolumn{2}{c}{\textbf{S (\%)}} & \multicolumn{2}{c}{\textbf{C (\%)}} & \multicolumn{2}{c}{\textbf{T (\%)}} & \multicolumn{2}{c}{\textbf{SD (m)}}
& \multicolumn{2}{c}{\textbf{S (\%)}} & \multicolumn{2}{c}{\textbf{C (\%)}} & \multicolumn{2}{c}{\textbf{T (\%)}} & \multicolumn{2}{c}{\textbf{SD (m)}}
& \multicolumn{2}{c}{\textbf{S (\%)}} & \multicolumn{2}{c}{\textbf{C (\%)}} & \multicolumn{2}{c}{\textbf{T (\%)}} & \multicolumn{2}{c}{\textbf{SD (m)}} \\
\cmidrule(lr){2-9} \cmidrule(lr){10-17} \cmidrule(lr){18-25}
\textbf{Setups} & \textbf{P} & \textbf{F} & \textbf{P} & \textbf{F} & \textbf{P} & \textbf{F} & \textbf{P} & \textbf{F} 
& \textbf{P} & \textbf{F} & \textbf{P} & \textbf{F} & \textbf{P} & \textbf{F} & \textbf{P} & \textbf{F} 
& \textbf{P} & \textbf{F} & \textbf{P} & \textbf{F} & \textbf{P} & \textbf{F} & \textbf{P} & \textbf{F} \\
\midrule
IQL (Gen)               & \textbf{100} & \textbf{96} & \textbf{0} & \textbf{1} & \textbf{0} & \textbf{3} & 3.89 & 4.40 & \textbf{98} & \textbf{93} & \textbf{2} & 5 & \textbf{0} & \textbf{2} & 3.75 & 4.07 & \textbf{98} & \textbf{91} & \textbf{2} & \textbf{3} & \textbf{0} & \textbf{6} & 3.82 & 4.29 \\
IQL (UDS)         & 99 & 92 & 1 & 2 & 0 & 6 & \textbf{4.07} & 4.12 & 97 & 85 & 2 & 6 & 1 & 9 & 3.83 & 4.08 & 95 & 89 & 3 & 0 & 2 & 11 & 3.90 & 4.08 \\
IQL (VLM)         & 99 & 80 & 1 & 2 & 0 & 18 & 3.95 & \textbf{4.53} & 93 & 87 & 7 & \textbf{1} & 0 & 12 & \textbf{4.02} & \textbf{4.55} & 95 & 77 & 3 & 6 & 2 & 17 & \textbf{3.96} & \textbf{4.66} \\
\midrule
CQL (Gen)               & 96 & \textbf{61} & \textbf{0} & \textbf{0} & 4 & \textbf{39} & 4.94 & 5.69 & \textbf{94} & \textbf{66} & 2 & 3 & \textbf{4} & \textbf{31} & \textbf{4.97} & \textbf{6.14} & 87 & 42 & 1 & \textbf{0} & 12 & 58 & \textbf{4.92} & 4.78 \\
CQL (UDS)               & 94 & 54 & 1 & 1 & 5 & 45 & \textbf{4.99} & \textbf{5.78} & 92 & 53 & \textbf{0} & \textbf{1} & 8 & 46 & 4.77 & 5.54 & 91 & 38 & \textbf{0} & 1 & 9 & 61 & 4.75 & \textbf{5.70} \\
CQL (VLM)               & \textbf{97} & \textbf{61} & 1 & \textbf{0} & \textbf{2} & \textbf{39} & 4.62 & 4.92 & \textbf{94} & 60 & 2 & 2 & 4 & 38 & 4.55 & 5.25 & \textbf{95} & \textbf{57} & 1 & 3 & \textbf{4} & \textbf{40} & 4.34 & 5.12 \\
\midrule
BPPO (Gen)               & \textbf{99} & \textbf{96} & \textbf{1} & \textbf{4} & \textbf{0} & \textbf{0} & \textbf{3.19} & \textbf{3.22} & \textbf{95} & \textbf{94} & \textbf{5} & \textbf{6} & \textbf{0} & \textbf{0} & 3.20 & \textbf{3.29} & 90 & \textbf{91} & 10 & \textbf{9} & \textbf{0} & \textbf{0} & \textbf{3.12} & 2.84 \\
BPPO (UDS)               & \textbf{99} & \textbf{96} & \textbf{1} & \textbf{4} & \textbf{0} & \textbf{0} & \textbf{3.16} & \textbf{3.06} & \textbf{95} & 90 & \textbf{5} & 10 & \textbf{0} & \textbf{0} & \textbf{3.83} & 2.92 & \textbf{91} & 89 & \textbf{9} & 11 & \textbf{0} & \textbf{0} & 2.70 & \textbf{2.92} \\
BPPO (VLM)               & 89 & 86 & 11 & 13 & \textbf{0} & 1 & 1.14 & 1.13 & 66 & 67 & 34 & 32 & \textbf{0} & 1 & 1.08 & 1.11 & 67 & 58 & 33 & 40 & \textbf{0} & 2 & 1.13 & 1.12 \\
\bottomrule
\end{tabular}
}
\end{table}

\subsection{Trajectory Analysis}
\begin{figure}[!h]
\centering
\includegraphics[width = 0.85 \linewidth]{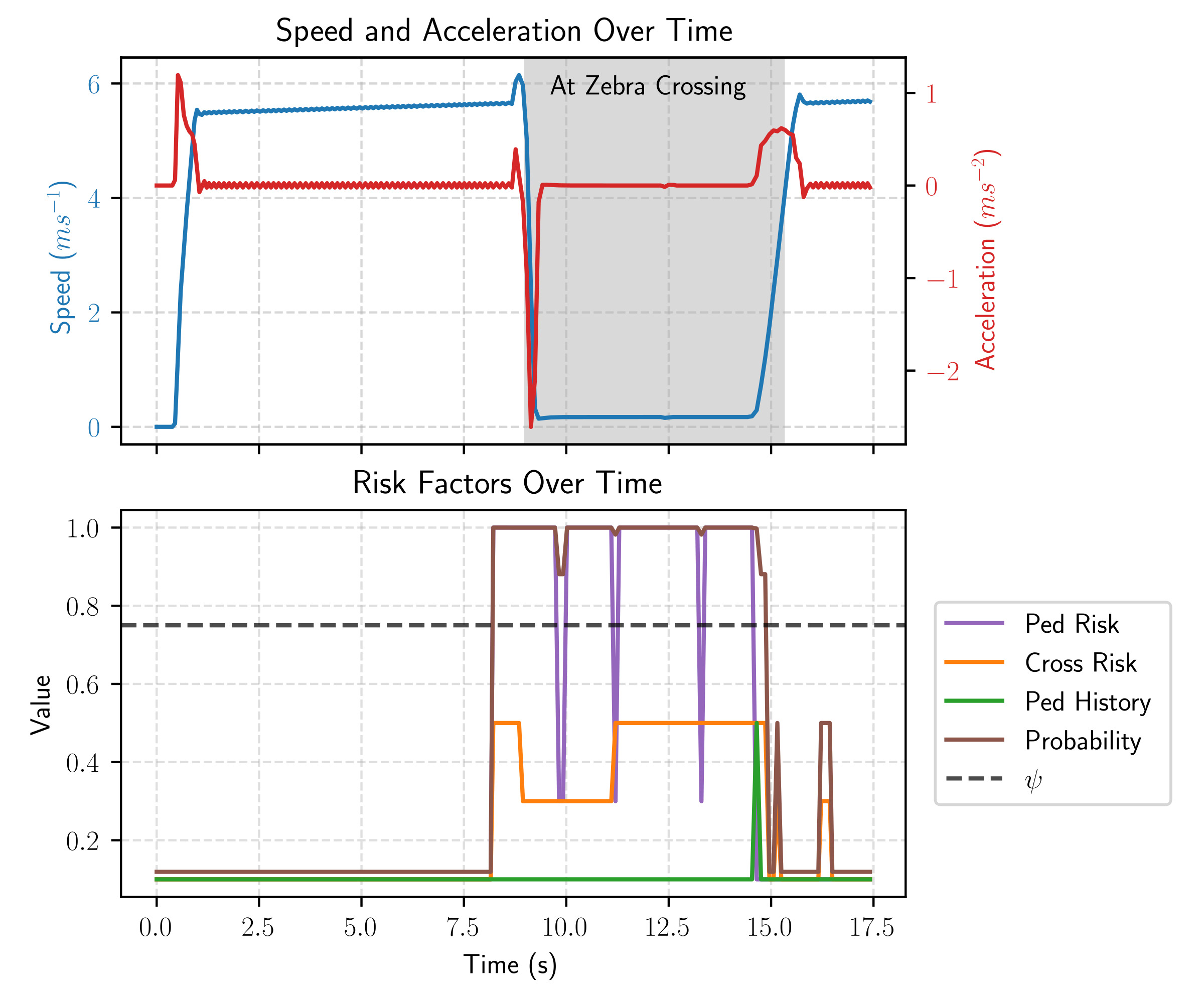}
\caption{Graphs showing a trajectory collected using the IQL (Gen) setup in the full-occlusion, high-pedestrian-density scenario variant. The top graph shows speed and acceleration over time; the grey-shaded area indicates the ego vehicle is at the zebra crossing. The bottom row illustrates how each risk factor used to determine $c_t$ changes throughout a trajectory. The black dotted line represents the threshold $\psi$; if the probability of stopping exceeds $\psi$, then $c_{t}=1$, and the agent prioritizes safety over efficiency in the reward function.}
\label{fig:kinematics}
\end{figure}

The final simulation-based experiment is a qualitative evaluation of the vehicle control quality. A trajectory was collected using the IQL (Gen) setup in the full-occlusion, high-pedestrian-density scenario variant. This scenario variant was selected because it proved most challenging in the collision test, and IQL (Gen) was chosen because it achieved the highest performance. The top row of Figure \ref{fig:kinematics} illustrates the speed and acceleration during the trajectory. To begin, the vehicle accelerates from a stationary position, driving at the speed limit until $t\approx9.0\text{s}$. From this point, the vehicle sharply decelerates as it approaches the zebra crossing, and remains stationary whilst the pedestrians cross the road. After $t\approx15\text{s}$, the vehicle accelerates again, driving at the speed limit, towards the goal point. Overall, the vehicle showed smooth control with minimal acceleration noise, and, most importantly, at the zebra crossing, it did not perform any false takeoffs, waiting until the zebra crossing was fully clear before accelerating. 

The final row in Figure \ref{fig:kinematics} illustrates the impact of the risk factors (defined in Section \ref{section:method_adaptive_safety_component}) on the probability of prioritizing safety or efficiency during a trajectory. There were minimal risk factors from $t=0\text{s}$ to $t\approx8\text{s}$ while the ego vehicle was driving. At this point, a zebra crossing is detected in the camera image, occluded by a van in the right lane, resulting in a zebra crossing risk of $0.5$. Simultaneously, the vehicle detects a pedestrian on the road, setting the pedestrian risk to $1$, the highest risk factor. The combined risk factors push the collision probability past the threshold $\psi=0.75$ (represented by the black dotted line), setting $c_t=1$ and prioritizing safety over efficiency. From $t\approx8\text{s}$ to $t\approx14.5\text{s}$, the vehicle remains stationary at the zebra crossing, with fluctuations in the pedestrian risk factor as pedestrians enter and leave the zebra crossing. Once the pedestrians leave the vehicle’s field of view, the risk factors decrease, lowering the probability to below $\psi$, resetting $c_t$ to $0$, and allowing the vehicle to prioritize efficiency and drive off. 

The following sub-section moves away from the experiments in CARLA, and presents the findings of applying the proposed reward labeling method to the subset of the A2D2 dataset detailed in Section \ref{section:real-world-application}.

\subsection{A2D2 Reward Labeling}
\begin{figure}[!th]
    \centering
    \begin{subfigure}[t]{0.48\textwidth}
        \centering
        \includegraphics[width=\linewidth]{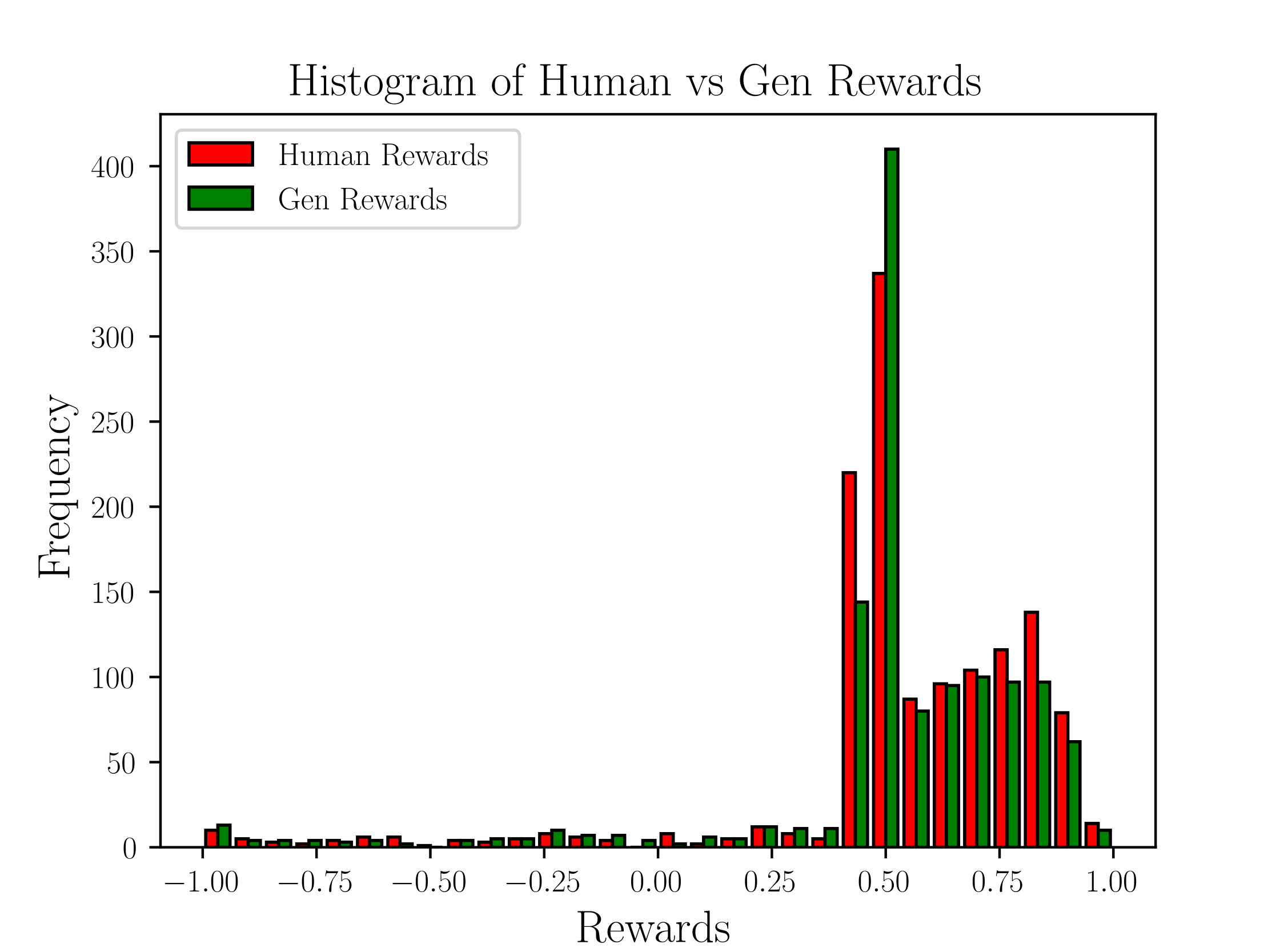}
        \caption{Histogram plot, with the distribution of Human Rewards in red and Gen Rewards in green.}
        \label{fig:a2d2_histogram}
    \end{subfigure}
    \hfill
    \begin{subfigure}[t]{0.48\textwidth}
        \centering
        \includegraphics[width=\linewidth]{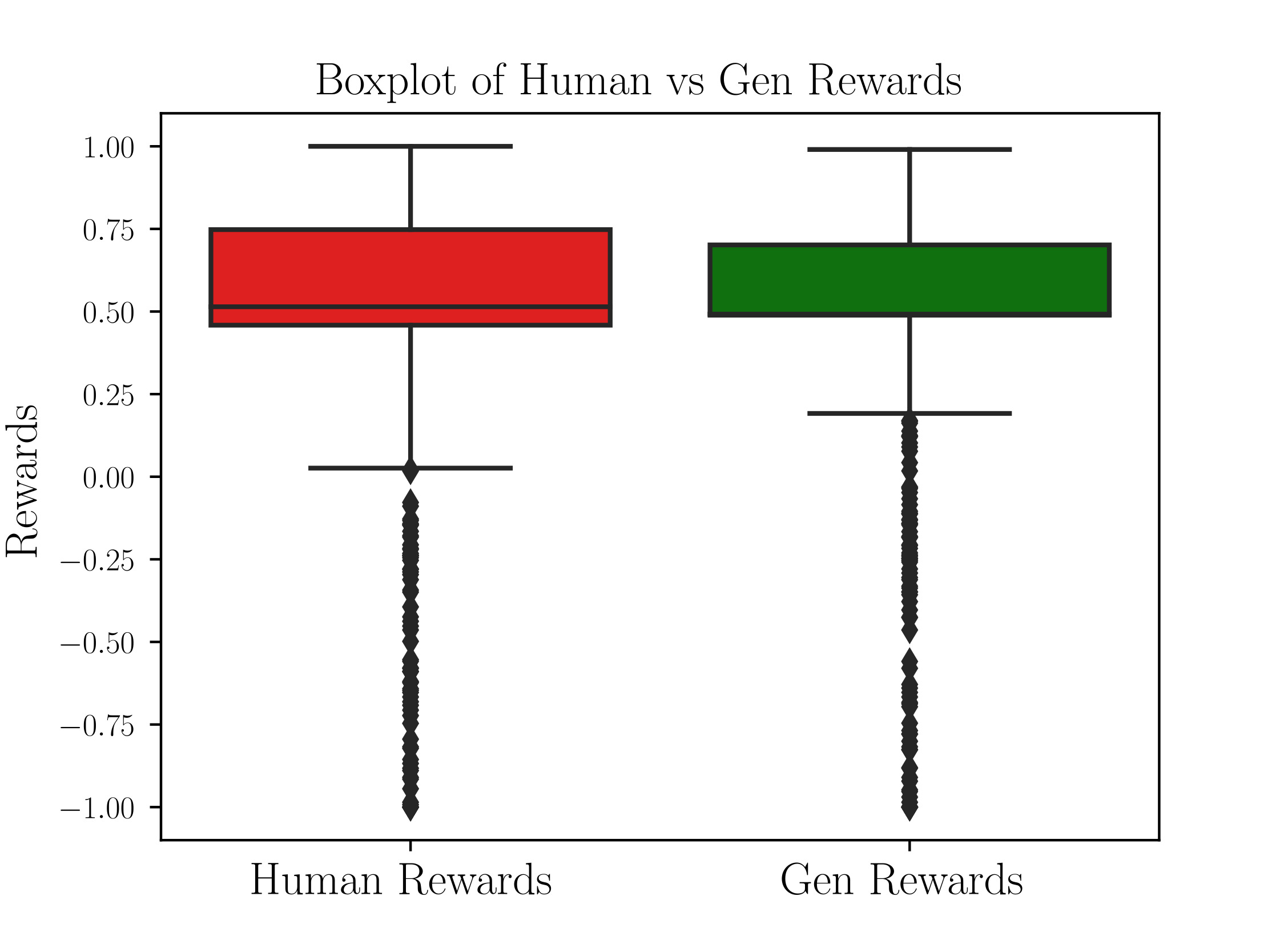}
        \caption{Box plot, with Human Rewards in red and Gen Rewards in green.}
        \label{fig:a2d2_boxplot}
    \end{subfigure}
    \caption{Histogram and boxplot illustrating the distribution of Human Rewards and Gen Rewards for a subset of the A2D2 dataset.}
    \label{fig:a2d2_dist_plot}
\end{figure}

Figure \ref{fig:a2d2_histogram} displays a histogram plot, showing the distribution of reward labels generated by the proposed method (referred to as "Gen Reward") and the human annotator (referred to as "Human Reward"). Overall, the two sets of reward labels have a similar distribution, with most values falling between $0.4$ and $0.9$. This trend was expected as the method flagged $341$ samples as unsafe ($c_t=1$), which was relatively close to the $318$ samples that the human annotator manually labeled as unsafe. Reward labels near $0$ indicate samples where the ego vehicle is driving slowly or is stationary. The larger negative reward labels indicate samples in which the method or the human annotator has flagged unsafe observations, and the ego vehicle is traveling at higher speeds. It should be noted that these larger negative rewards are the least frequent, resulting in a long tail for both distributions. Finally, the reward labels above $0$ relate to instances where the ego vehicle is driving under safe observations.

Figure \ref{fig:a2d2_boxplot} illustrates a box plot for the two distributions of reward labels, and the main takeaways complement the findings from Figure \ref{fig:a2d2_histogram}. Both distributions have similar medians near $0.5$. However, Human Rewards have a slightly higher interquartile range (IQR) than Gen Rewards, suggesting that the human annotations exhibit slightly more variability in positive values. Additionally, Gen Rewards has a tighter value range than Human Rewards and a wider outlier range, suggesting that the proposed method has assigned larger negative rewards more frequently to cases it has labeled as unsafe.

\begin{figure}[!th]
\centering
\includegraphics[width = 0.75 \linewidth]{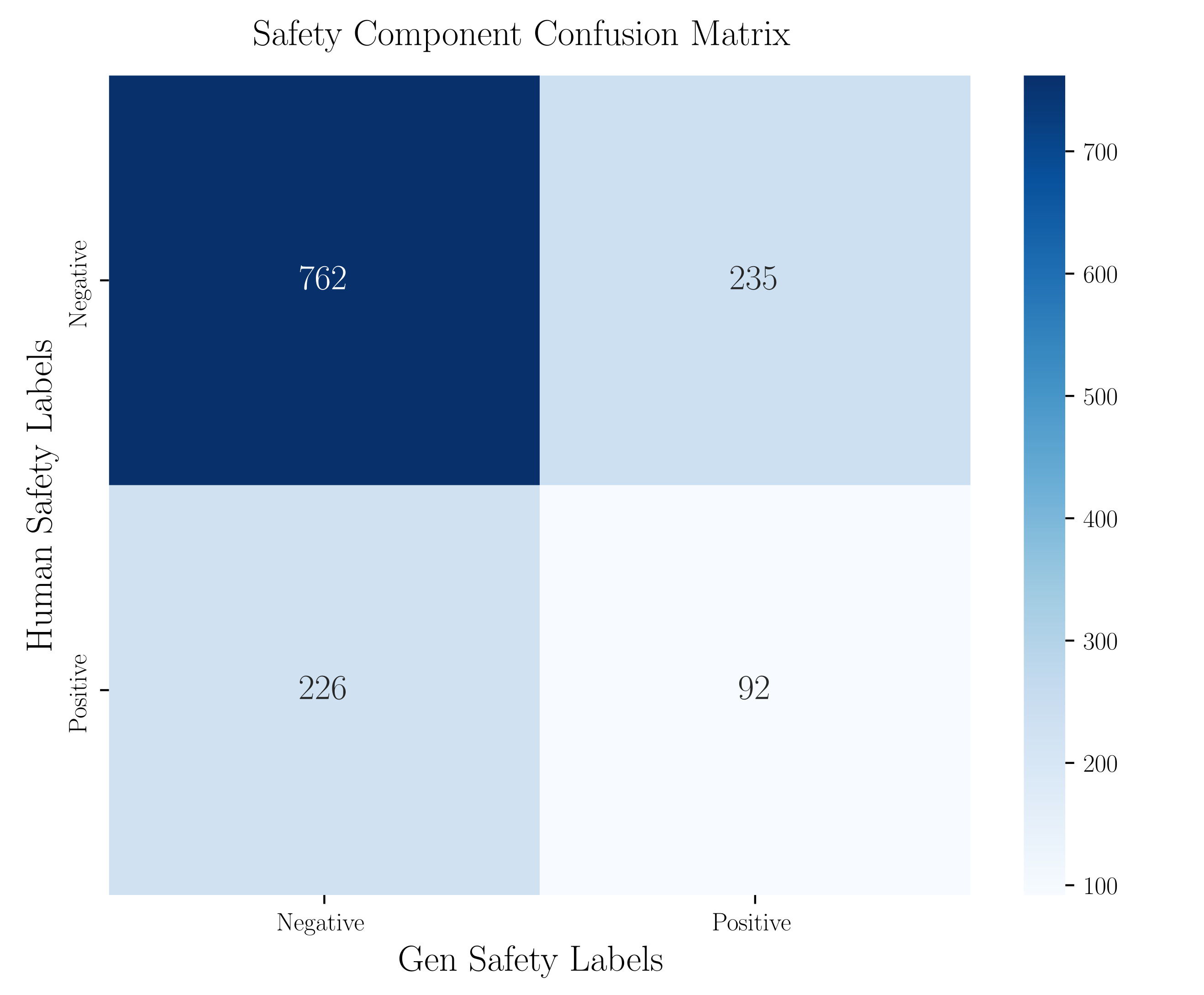}
\caption{Confusion matrix showing the agreement and disagreements between human-assigned safety labels and those predicted by the pipeline.}
\label{fig:confusion_matrix}
\end{figure}

Figure \ref{fig:a2d2_dist_plot} indicates that Gen Rewards and Human Rewards have similar distributions, and this is attributed to the same reward function (Equation \ref{eqn:final_reward}) being used. What distinguishes the two reward sets is how the adaptive safety component (ct) is derived. Figure \ref{fig:confusion_matrix} shows a confusion matrix of the safety components produced by the two methods, where a discrepancy between the two component sets is revealed. Specifically, of the $318$ samples the human annotator labeled as unsafe, the proposed method agreed with the annotator on only $28.9\%$ of samples, labeling the rest as safe. Additionally, when examining the samples the human annotator labeled as safe, the method aligned with the human annotator on $762$ samples, whereas it opposed the human annotator on $235$ samples. An initial review of the results suggests that the method struggles to align with human judgment in classifying observations as safe or unsafe. However, a deeper analysis of the instances where the method and human annotator disagree reveals a more nuanced issue. 
\begin{figure}[!ht]
    \centering
    \fbox{%
    \begin{minipage}{\textwidth}
        \centering
        \begin{subfigure}[t]{0.48\textwidth}
            \centering
            \includegraphics[width=\linewidth]{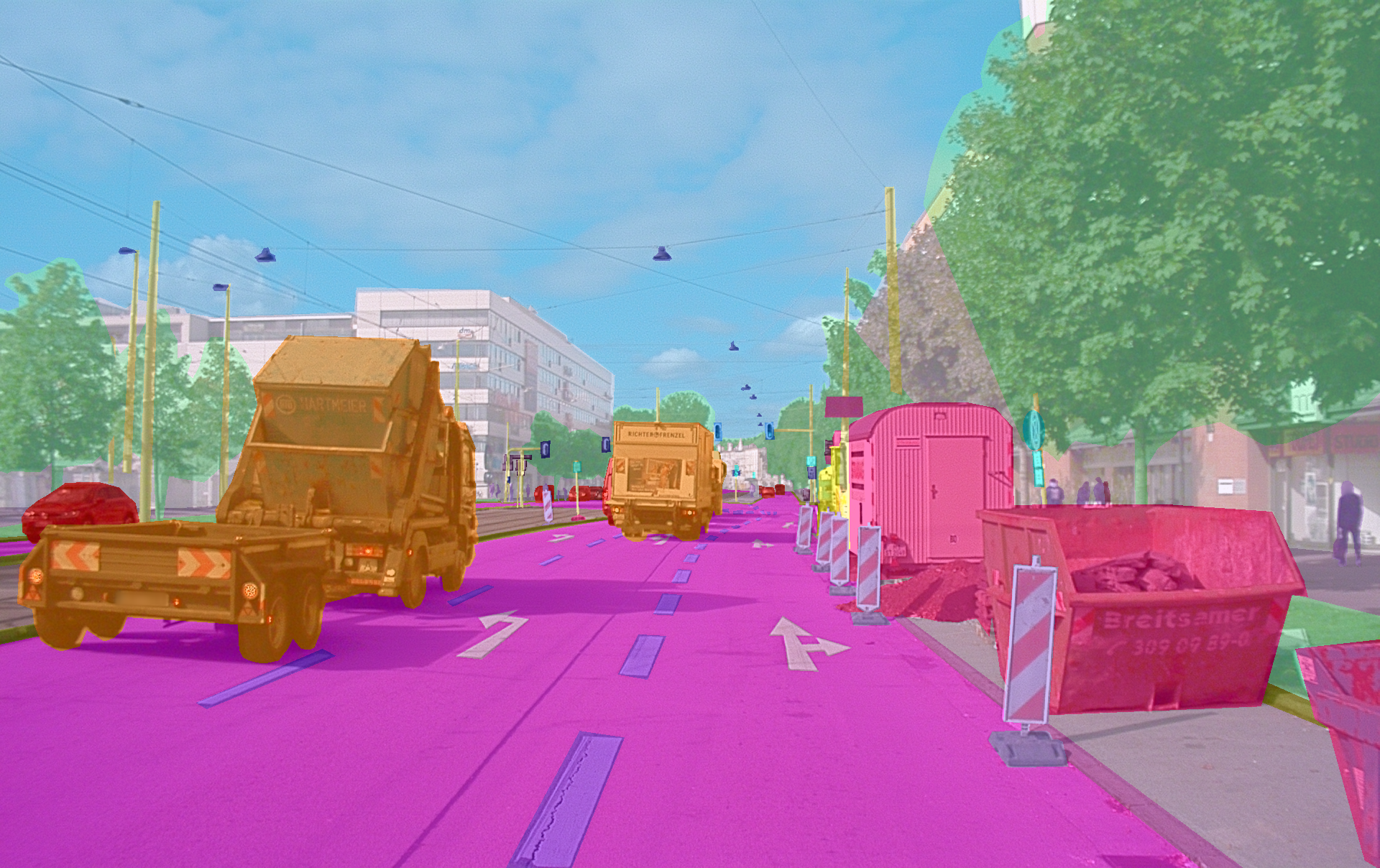}
            \caption{\small Both the proposed method and human annotator labeled the observation as safe ($c_t=0$). The ego vehicle is driving in a clear lane, with no pedestrians in view.}
            \label{fig:a2d2_p0h0}
        \end{subfigure}
        \hfill
        \begin{subfigure}[t]{0.48\textwidth}
            \centering
            \includegraphics[width=\linewidth]{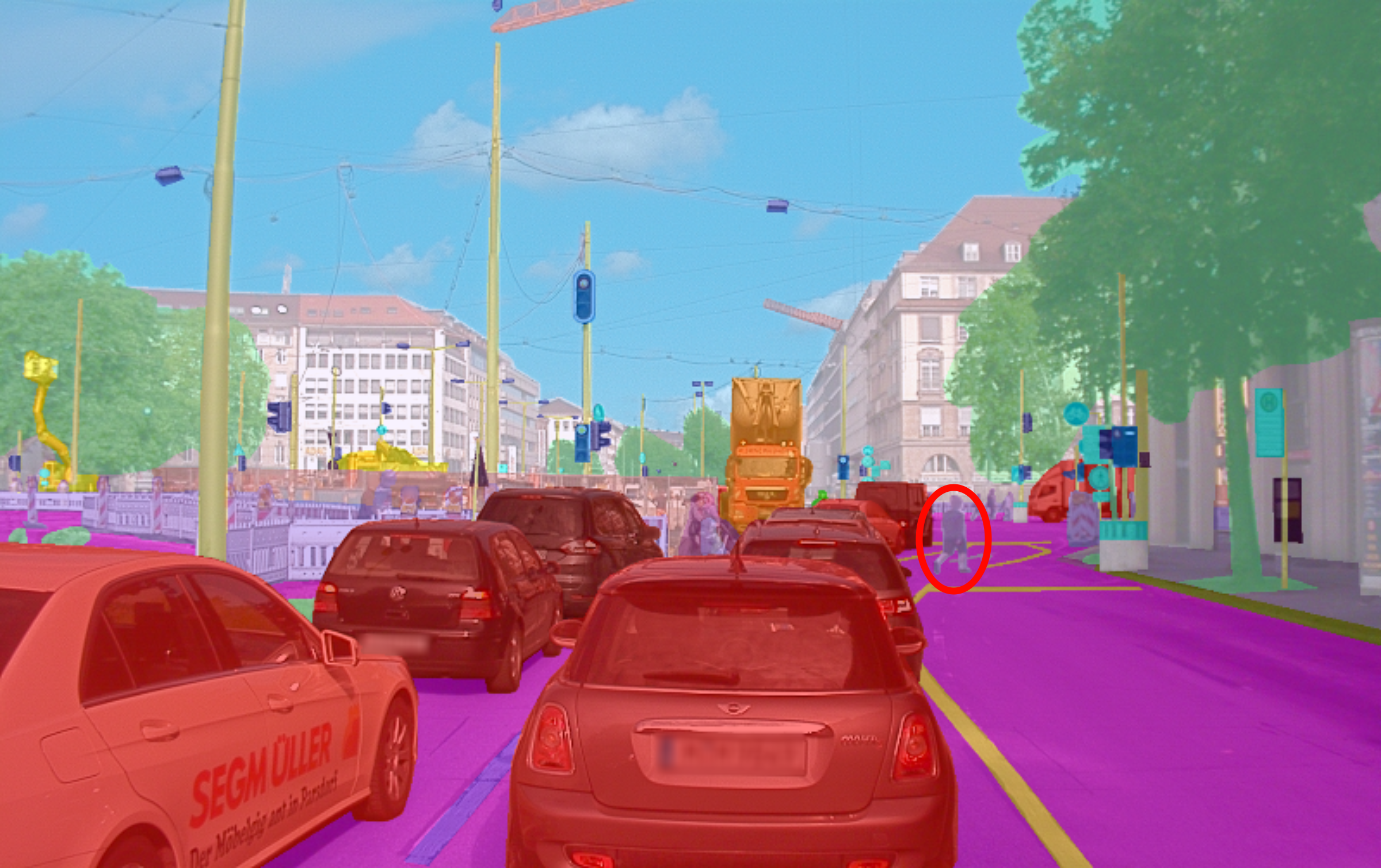}
            \caption{\small Human annotator labeled the sample as unsafe ($c_t=1$), while the proposed method labeled it as safe ($c_t=0$). The ego vehicle is in traffic, whilst a pedestrian crosses the road (circled in red).}
            \label{fig:a2d2_p0h1}
        \end{subfigure}
        \vspace{1em}
        \begin{subfigure}[t]{0.48\textwidth}
            \centering
            \includegraphics[width=\linewidth]{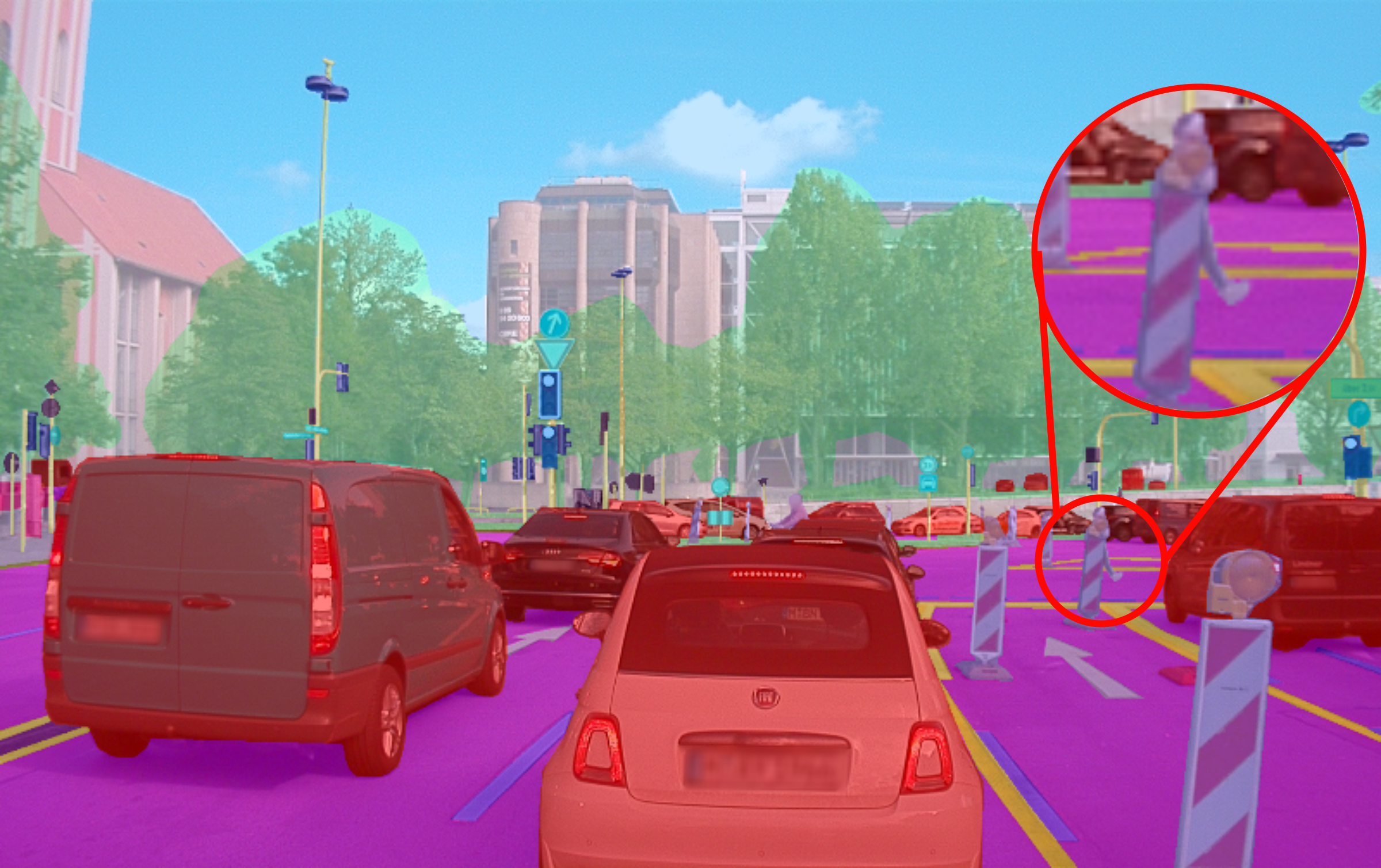}
            \caption{\small Human annotator labeled the observation as safe ($c_t=0$), while the proposed method classified it as unsafe ($c_t=1$). The ego vehicle is stationary in traffic, and a pedestrian, heavily occluded by a road sign, is crossing the road (circled in red).}
            \label{fig:a2d2_p1h0}
        \end{subfigure}
        \hfill
        \begin{subfigure}[t]{0.48\textwidth}
            \centering
            \includegraphics[width=\linewidth]{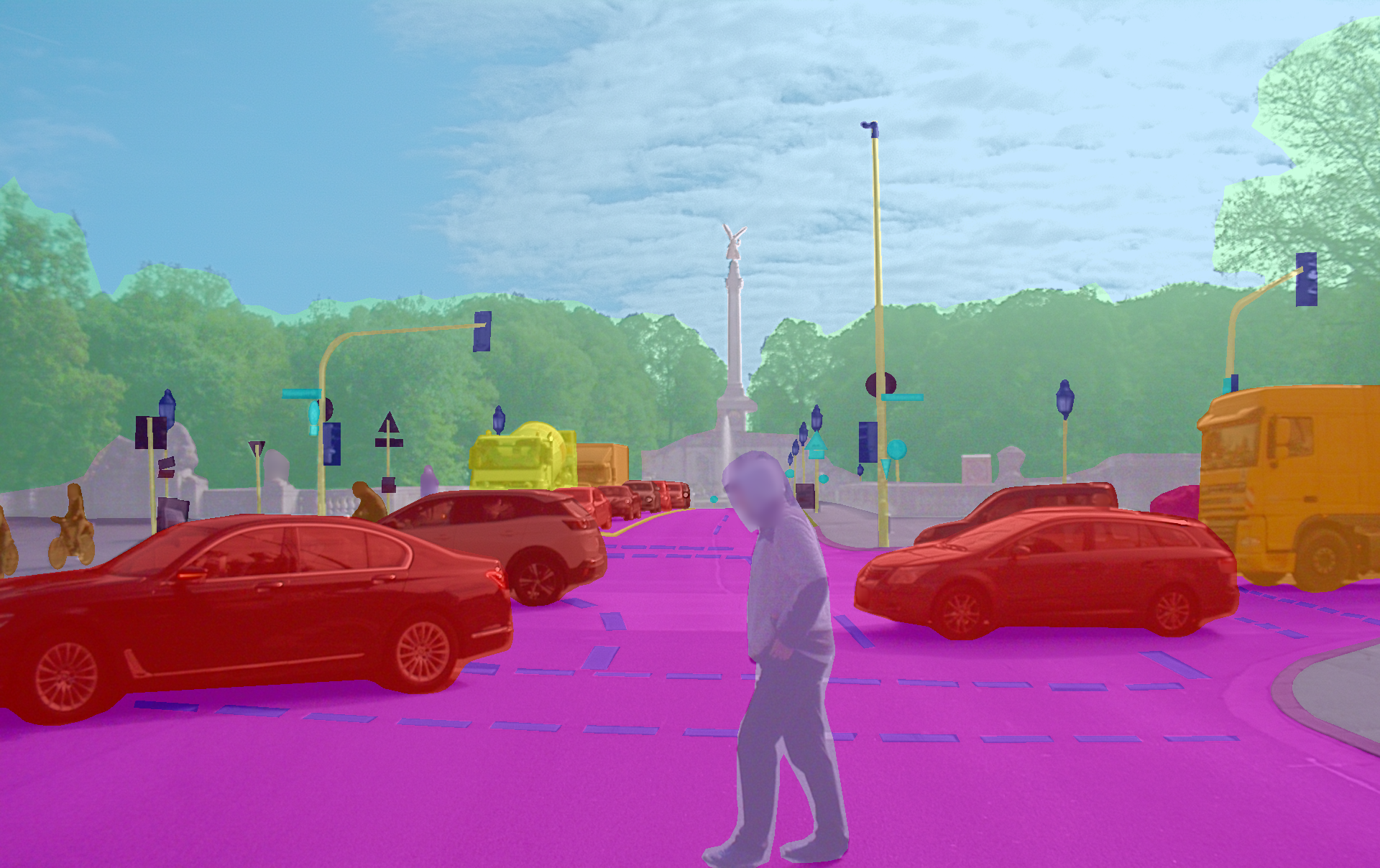}
            \caption{\small Both the proposed method and human annotator labeled the observation as unsafe ($c_t=1$). The ego vehicle is stationary at a junction and a pedestrian is directly in front of the ego vehicle, crossing the road.}
            \label{fig:a2d2_p1h1}
        \end{subfigure}
    \end{minipage}
    }
    \caption{\small A2D2 image-semantic map overlay examples of the proposed method and human annotator assigning safety labels.}
    \label{fig:a2d2_labeling_examples}
\end{figure}

Figure \ref{fig:a2d2_labeling_examples} illustrates instances where the proposed method and human annotator either agreed or disagreed on labeling observations as safe or unsafe. While Figures \ref{fig:a2d2_p0h0} and \ref{fig:a2d2_p1h1} show cases of agreement, the more striking result is seen in Figure \ref{fig:a2d2_p1h0}, where the human annotator labeled the sample as safe, missing a heavily occluded pedestrian crossing the road. In contrast, the proposed method correctly identified this potential hazard, leveraging the semantic segmentation map for more detailed analysis. However, this capability is inconsistent, as shown in Figure \ref{fig:a2d2_p0h1}, where the proposed method fails to recognize a clearly visible pedestrian crossing the road, labeling the sample as safe. This mislabeling may be linked to limitations in the risk factors used to determine the safety component ($c_t$). Specifically, the pedestrian risk factor in Figure \ref{fig:risk_factors} assesses the pedestrian's position by assigning a higher risk if the $10$ pixels below the pedestrian are in the road class. In Figure \ref{fig:a2d2_p0h1}, the pixels beneath the pedestrian include both road and road markings, which may have confused the proposed method. Moreover, this reveals a notable limitation in the method's capacity to effectively capture the spatial relationships between actors within a scene. Evaluating a pedestrian as a hazard requires considering important contextual factors, such as their direction and speed, in addition to their position. The current method does not include this information in the decision-making process. 

\section{Discussion and Future Works}
\label{section:discussion}
The collision test highlights a significant limitation of the proposed pipeline: its performance degrades in highly congested environments. This is understandable, given that the current pipeline relies solely on front-facing image observations with limited temporal context. The pipeline could be enhanced by integrating multiple camera feeds to overcome these limitations, providing a more comprehensive representation of the vehicle’s surroundings. Additionally, the semantic maps from different viewpoints could be used to build a more meaningful, symbolic representation of objects in the driving scene \cite{neubert2021vector}. Furthermore, the adaptive component of the reward function, $c_t$, could be improved by incorporating temporal context through a more sophisticated feature extractor like video transformers and a memory bank designed to isolate key social cues, such as pedestrian positions and their relevance in the current scene \cite{marchetti2024smemo}. 

Regarding the use of VLM for reward labeling, the simulation-based experiments did not provide enough evidence to recommend the method. In addition to the considerably higher computational load of VLMs, the collision test highlighted another issue: using VLMs to produce dense reward signals is inherently a black-box approach. If the VLM rewards lead to a sub-optimal policy, it can be challenging to diagnose the system and understand the rationale of the reward labels. The current robotics literature instead uses VLMs to label trajectories for sparse-reward tasks or preference-based RL~\cite{venkataraman2024real}, \cite{wang2025prefclm}, and has seen success. It is recommended to use VLMs in this context rather than producing per-sample reward values.

While the A2D2 reward labeling experiment highlighted a significant discrepancy between the pipeline and a human annotator in identifying safe and unsafe samples, the results are inconclusive due to limited data and reliance on a single annotator. To draw more robust conclusions about the pipeline's effectiveness, it is crucial to involve a larger, more diverse group of annotators with varying levels of driving experience, as perceptions of safety can vary widely. Additionally, the definition of `safe' must be reconsidered, as what is deemed safe for the driver or passenger of the ego vehicle may not be perceived the same way by other road users, such as pedestrians or cyclists. Future work should focus on establishing a comprehensive and inclusive method for safety assessment that accounts for the perspectives of all road users.

Figures \ref{fig:confusion_matrix} and \ref{fig:a2d2_labeling_examples} from the A2D2 reward labeling experiment expose a critical issue: the absence of robust metrics to effectively compare a model’s decision-making process with human reasoning. In online settings, model performance can be evaluated using tangible metrics such as success rate and quality of vehicle control. However, without interacting the environment, it is challenging to assess how well a model emulates human-like decision-making. Current methods, which rely on matching predicted outputs with human labels, are too simplistic and fail to capture the nuances of human judgment. This highlights the urgent need for more sophisticated evaluation criteria that can better quantify a model’s alignment with complex human reasoning, especially in scenarios where direct interaction and feedback are not possible.

To fully assess the quality of the reward labels generated for the A2D2 dataset, a policy trained on this labeled data needs to be evaluated. However, obtaining real-world results from such evaluations requires significant resources, making it impractical in many cases. An alternative approach is Offline Policy Evaluation (OPE), which could allow for a more resource-efficient assessment. Unfortunately, OPE remains a persistent challenge in Offline RL, with minimal progress in recent years~\cite{levine2020offline}, \cite{prudencio2023survey}. A robust OPE technique would be invaluable for evaluating policies during training, ensuring their effectiveness and safety before deployment in real-world environments.

\section{Conclusion}
\label{section:conclusion}
In conclusion, a method was presented to generate human-aligned reward labels for real-world AD datasets, enabling Offline RL. The reward function includes an adaptive safety component, allowing the agent to prioritize safety over efficiency in potential collision scenarios. The viability of the labels was demonstrated in an occluded pedestrian crossing scenario with varying pedestrian traffic in CARLA, using the various Offline RL algorithms for longitudinal control. The method was compared with a simpler baseline that leverages unlabeled data and a VLM method. The simulation-based results showed that the method could produce meaningful rewards; however, success rates in the collision tests decreased with higher pedestrian density. The method was also applied to a subset of the A2D2 dataset, focusing on pedestrian crossings. The generated rewards were compared with a human annotator, who manually labeled samples deemed unsafe, and the comparison showed a noticeable disparity between the two sets of reward labels. Most interestingly, in some cases, the method flagged data samples as unsafe that the human annotator had missed. However, due to a lack of policy training and real-world evaluation, the quality of the reward labels was not fully guaranteed.

Directions for future work include incorporating  semantic maps from multiple camera feeds to build a more comprehensive and symbolic representation of the ego vehicle's surroundings, enabling better performance in dense traffic settings. In addition, to improve the pipeline's temporal reasoning, a more sophisticated feature extractor, such as a video transformer, can be used alongside a socially aware memory bank.





\bibliographystyle{elsarticle-num} 
\bibliography{ref}

\end{document}